\documentclass[10pt,twocolumn,letterpaper]{article}

\usepackage{cvpr}              

\definecolor{cvprblue}{rgb}{0.21,0.49,0.74}
\usepackage{graphicx} 
\usepackage{subcaption}
\usepackage[pagebackref,breaklinks,colorlinks,allcolors=cvprblue]{hyperref}
\usepackage{multirow}
\usepackage{graphicx}
\usepackage{booktabs}
\usepackage{array}
\usepackage{makecell}
\def\paperID{7} 
\def\confName{CVPR}
\def\confYear{2026}

\title{Can Nano Banana 2 Replace Traditional Image Restoration Models? \\
An Evaluation of Its Performance on Image Restoration Tasks}

\author{Weixiong Sun$^1$, Xiang Yin$^2$, Chao Dong$^{31}$\\
\small $^1$Shenzhen University of Advanced Technology \quad $^2$Fudan University \\
\small $^3$Shenzhen Institutes of Advanced Technology, Chinese Academy of Sciences
}

\begin{document}
\maketitle
\begin{abstract}
Recent advances in generative AI raise the question of whether general-purpose image editing models can serve as unified solutions for image restoration. We conduct a systematic evaluation of Nano Banana 2 across diverse scenes and degradations. Our results show that prompt design is critical, with concise prompts and explicit fidelity constraints achieving a better balance between reconstruction and perceptual quality. 
Nano Banana 2 achieves competitive full-reference performance and is consistently preferred in user studies, while showing strong generalization in challenging scenarios. However, we observe a gap between perceptual quality and restoration fidelity, as the model tends to produce visually rich results with over-enhanced details and inconsistencies. This issue is not well captured by existing IQA metrics or user studies. 
Overall, general-purpose models show promise as unified IR solvers from a perceptual perspective, but require improved controllability and fidelity-aware evaluation.
Further comparisons and detailed analyses are available in our project repository: \url{https://github.com/yxyuanxiao/NanoBanana2TestOnIR}.

\end{abstract}    
\vspace{-18pt}
\section{Introduction}
\label{sec:intro}
\vspace{-8pt}
The rapid evolution of generative AI has fundamentally reshaped the landscape of computer vision, with recent general-purpose image generation and editing models~\cite{rombach2022high, saharia2022photorealistic, brooks2023instructpix2pix, xai_grok_imagine_image_2026, cao2025hunyuanimage, seedream2025seedream} demonstrating remarkable capabilities in high-fidelity content synthesis, semantic manipulation, and instruction-following~\cite{betker2023improving, podell2023sdxl, seedream2025seedream}. 
At the same time, image restoration (IR) remains a core problem in low-level vision, encompassing tasks such as denoising~\cite{zhang2017beyond}, deblurring~\cite{kupyn2018deblurgan}, super-resolution~\cite{dong2015image, wang2018esrgan}, and artifact removal~\cite{dong2015compression}. 
Although substantial progress has been made in these tasks, especially with the recent success of diffusion-based restoration models~\cite{rombach2022high, wang2024sinsr, fei2023generative, wang2024exploiting}, existing approaches are still largely developed within a restoration-specific paradigm~\cite{zhou2025unires, he2025diffusion, jiang2025survey}, where they are typically designed for predefined degradation types, optimized for relatively closed distributions, and trained to solve one or one family of degradations at a time~\cite{song2019generative, song2020denoising}. 
As a result, their effectiveness degrades in real-world scenarios, where degradations are mixed, spatially non-uniform and frequently unknown a priori\cite{wang2024exploiting, zhou2025unires}.

This limitation motivates a broader question: can a general-purpose image editing model~\cite{team2023gemini, labs2025flux, cai2025z, wu2025qwen} serve as a unified solver for image restoration?
Compared with specialist restoration models, such a model is appealing. 
First, it may provide a single framework capable of handling diverse degradations without requiring task-specific retraining. 
Second, it can potentially exploit large-scale semantic and structural priors acquired during pretraining, enabling plausible reconstruction in severely ill-posed cases where low-level information is heavily corrupted or even missing. 
Third, it naturally supports instruction-driven control, allowing restoration to be conditioned not only on degraded observations, but also on user intent.
These properties suggest a fundamentally different restoration paradigm, one that moves beyond deterministic signal recovery toward a more general process of semantic reconstruction and controllable visual enhancement.

\begin{table*}[tp]
\centering
\footnotesize
\begin{tabular}{>{\centering\arraybackslash}p{1.45cm}|>{\centering\arraybackslash}p{0.5cm}|p{14.1cm}}
\hline
\textbf{Category} & \textbf{ID} & \multicolumn{1}{c}{\textbf{Prompt}} \\
\hline

\multirow[c]{6}{*}{\makecell{Short \\ \scriptsize (29--30 words)}}
& \multirow[c]{2}{*}{S1} & Restore and enhance the image by removing noise, compression artifacts, and blur while improving sharpness and natural colors. If people appear in the scene, recover clear facial features, skin texture. \\
& \multirow[c]{2}{*}{S2} & Improve the image quality by correcting blur, noise, and compression artifacts and enhancing clarity and color fidelity. If the image contains people, restore facial details and natural skin textures. \\
& \multirow[c]{2}{*}{S3} & Enhance and restore the image to produce a clean and detailed photograph with improved sharpness and colors. If humans are present, reconstruct clear facial features and realistic skin tones. \\
\hline

\multirow[c]{6}{*}{\makecell{Short +\\ Fidelity \\ \scriptsize (26--28 words)}}
& \multirow[c]{2}{*}{SF1} & Restore the image by reducing noise and blur while preserving the original content and scene structure. If people appear, improve facial clarity while keeping their identity unchanged. \\
& \multirow[c]{2}{*}{SF2} & Enhance the image quality by removing degradations while maintaining the original visual fidelity and scene details. If humans are present, restore facial features without altering identity. \\
& \multirow[c]{2}{*}{SF3} & Improve clarity and remove noise and compression artifacts while preserving the original structure and realism of the image. If people appear, enhance facial details while keeping them consistent.\\
\hline

\multirow[c]{8}{*}{\makecell{Long \\ \scriptsize (35--47 words)}}
& \multirow[c]{3}{*}{L1} & Restore and enhance the image by addressing common degradations such as noise, compression artifacts, and blur. Improve sharpness, clarity, and natural colors to produce a clean and detailed photograph. If the scene contains people, restore high-quality facial features, clear eyes, natural skin texture, and realistic skin tones.\\
& \multirow[c]{2}{*}{L2} & Enhance and restore the image by removing noise, blur, and compression artifacts while improving fine details, sharpness, and color fidelity. If people are present, carefully reconstruct facial details, eyes, hair texture, and natural skin appearance.\\
& \multirow[c]{3}{*}{L3} & Improve the overall visual quality of the image by repairing degradations and enhancing clarity, texture, and natural colors. If the image includes people, restore sharp facial structures, detailed eyes, and realistic skin textures while keeping their appearance natural.\\
\hline

\multirow[c]{7}{*}{\makecell{Long +\\ Fidelity \\ \scriptsize (34--38 words)}}
& \multirow[c]{3}{*}{LF1} & Restore and enhance the image by removing noise, blur and compression artifacts while strictly preserving the original scene structure and visual fidelity. If the scene contains people, improve facial clarity and skin texture while keeping their identity unchanged. \\

& \multirow[c]{2}{*}{LF2} & Enhance the image quality while maintaining high fidelity to the original content and avoiding changes to scene structure. If people are present, restore facial features and skin details while preserving their identity and natural appearance. \\

& \multirow[c]{2}{*}{LF3} & Restore the degraded image by improving clarity, reducing noise and blur, and enhancing colors while preserving the original visual content. If the image includes people, refine facial details without altering their identity or expression. \\

\hline
\end{tabular}
\vspace{-8pt}
\caption{A taxonomy of prompt designs used in our experiments, categorized by length and the inclusion of fidelity constraints.}
\label{tab:prompts}
\vspace{-19pt}
\end{table*}

However, this promise also introduces a fundamental tension between perceptual plausibility and pixel-level fidelity~\cite{blau2018perception}.
Strong generative priors can produce outputs that appear sharper, cleaner, and more visually appealing to human observers, yet still deviate substantially from the ground-truth image. 
This tension becomes especially important in image restoration, where success is not only determined by perceptual quality, but also by fidelity to the original content.
Recent analyses~\cite{yin2026far, zuo2025nano} further suggest that such models may excel at perceptual enhancement while remaining vulnerable to hallucinated textures, semantic drift, structural inconsistencies, and color deviations, failing to achieve the requirements of IR.

The recent release of Nano Banana 2~\cite{team2023gemini} makes this question particularly timely. 
As a newly introduced high-profile general IR and editing model, Nano Banana 2 has quickly attracted broad attention, raising immediate interest in whether its capabilities extend beyond creative synthesis into traditional restoration problems. 
Yet, despite this attention, its behavior on image restoration tasks remains largely underexplored. 
It is still unclear whether Nano Banana 2 can function as a reliable unified restorer, whether its strong generative prior improves restoration under severe or mixed degradations, and where it fails due to hallucination, semantic alteration, or lack of fidelity constraints. 
These open questions bear on a more fundamental issue for the community: are we approaching a stage where powerful generalist image models can subsume specialized low-level vision systems, or do restoration tasks still require domain-specific designs and constraints?

In this work, we conduct a systematic study of Nano Banana 2 on image restoration tasks. Our goal is to assess whether a highly capable general-purpose image editing model can act as a credible unified restorer under open-world degradations. 
More specifically, we investigate how different prompting strategies affect restoration performance, how stable the model remains under prompt variations, how it behaves across diverse scenes and degradation types, and where it stands relative to state-of-the-art (SOTA) image restoration models. 
Through this study, we seek to clarify the true boundary between general image editing and reliable image restoration, and to provide empirical evidence for future restoration frameworks that may combine the perceptual power of generative models with the fidelity guarantees of specialist approaches.

\vspace{-12pt}
\section{Related Work}
\vspace{-6pt}
\subsection{Image Restoration}
\vspace{-5pt}

Image restoration is a core problem in low-level vision, aiming to recover high-quality images from degraded observations~\cite{jinjin2020pipal, zhang2022accurate}. 
Early studies mainly relied on handcrafted priors and optimization-based formulations~\cite{dabov2007image, tomasi1998bilateral}. 
With the rise of deep learning, data-driven restoration models, including convolutional~\cite{dong2015image, dong2016accelerating, wang2018esrgan} and transformer-based~\cite{zamir2022restormer, liang2021swinir} architectures, have achieved strong performance on a wide range of IR tasks.

Recently, diffusion-based models have emerged as a powerful paradigm for image restoration~\cite{rombach2022high, wang2024sinsr, fei2023generative, wang2024exploiting}. 
By leveraging strong generative priors learned from large-scale data, these methods can produce realistic, high-quality outputs. 
Representative approaches~\cite{yu2024scaling} combine restoration objectives with diffusion processes to improve both fidelity and perceptual quality. 
Despite their impressive performance, these methods still largely operate within restoration-specific frameworks~\cite{zhou2025unires, he2025diffusion, jiang2025survey} often relying on carefully designed pipelines, task-aware conditioning, or specialized training procedures~\cite{yu2024scaling, yu2025unicon}.

\vspace{-7pt}
\subsection{Image Editing Models}
\vspace{-4pt}
Recent advances in generative AI have led to a new class of general-purpose image editing models~\cite{team2023gemini, labs2025flux, cai2025z} that support image-to-image transformation through text instructions and visual conditions. 
Unlike traditional low-level vision systems, these models are built upon large-scale generative priors and are designed primarily for flexible content manipulation, controllable synthesis, and semantic-level visual editing. 
These models have demonstrated strong capabilities in producing high-quality and visually coherent outputs across diverse image editing scenarios.

A key property of these models is that they treat image transformation as a conditional generation problem, rather than a task-specific restoration process. 
This enables them to perform a wide range of edits within a unified framework and to respond naturally to language-based instructions~\cite{zhao2024ultraedit}. 
However, such flexibility also introduces uncertainty when they are applied to image restoration. 
Since these models are not explicitly optimized for recovering degraded observations, it remains unclear whether they can faithfully preserve source content under degradation, or whether their strong generative priors instead lead to semantic drift, hallucinated details, or structural deviations~\cite{cohen2024looks}.

Therefore, although general-purpose image editing models have shown impressive performance in creative and instruction-guided visual tasks, their capability boundary in image restoration remains largely underexplored. 
Our work focuses on this gap by systematically evaluating whether Nano Banana 2, as a representative high-profile general image editing model, can function as a reliable restorer under diverse restoration scenarios.

\begin{table}[tp]
\centering
\small
\setlength{\tabcolsep}{2pt}
\renewcommand{\arraystretch}{0.95}
\resizebox{\columnwidth}{!}{
\begin{tabular}{c|c|cccccc}
\toprule
Group & Prompt & PSNR & SSIM & LPIPS$\downarrow$ & MUSIQ & MANIQA & CLIP-IQA \\
\midrule

\multirow{4}{*}{L}
& L1 & 22.910 & 0.670 & 0.223 & 69.339 & 0.393 & 0.667 \\
& L2 & 23.211 & 0.668 & 0.222 & 69.462 & 0.401 & 0.684 \\
& L3 & 22.729 & 0.666 & 0.221 & 69.255 & 0.394 & 0.677 \\
\cmidrule(lr){2-8}
& \textbf{Avg} & \textbf{22.950} & \textbf{0.668} & \textbf{0.222} & \textbf{69.352} & \textbf{0.396} & \textbf{0.676} \\
\midrule

\multirow{4}{*}{LF}
& LF1 & 23.284 & 0.671 & 0.216 & 67.365 & 0.374 & 0.659 \\
& LF2 & 23.018 & 0.665 & 0.223 & 68.386 & 0.382 & 0.667 \\
& LF3 & 22.265 & 0.662 & 0.223 & 68.170 & 0.380 & 0.666 \\
\cmidrule(lr){2-8}
& \textbf{Avg} & \textbf{22.856} & \textbf{0.666} & \textbf{0.221} & \textbf{67.974} & \textbf{0.379} & \textbf{0.664} \\
\midrule

\multicolumn{2}{c|}{\textbf{Avg (L + LF)}} 
& \textbf{22.903} & \textbf{0.667} & \textbf{0.221} & \textbf{68.663} & \textbf{0.387} & \textbf{0.670} \\
\midrule

\multirow{4}{*}{S}
& S1 & 22.920 & 0.664 & 0.224 & 68.809 & 0.386 & 0.669 \\
& S2 & 22.655 & 0.665 & 0.224 & 68.768 & 0.387 & 0.670 \\
& S3 & 21.975 & 0.655 & 0.237 & 70.444 & 0.408 & 0.686 \\
\cmidrule(lr){2-8}
& \textbf{Avg} & \textbf{22.516} & \textbf{0.662} & \textbf{0.228} & \textbf{69.340} & \textbf{0.394} & \textbf{0.675} \\
\midrule

\multirow{4}{*}{SF}
& SF1 & 23.003 & 0.671 & 0.215 & 67.496 & 0.371 & 0.650 \\
& SF2 & 23.090 & 0.663 & 0.217 & 66.748 & 0.374 & 0.655 \\
& SF3 & 23.360 & 0.672 & 0.215 & 67.794 & 0.377 & 0.665 \\
\cmidrule(lr){2-8}
& \textbf{Avg} & \textbf{23.151} & \textbf{0.669} & \textbf{0.216} & \textbf{67.346} & \textbf{0.374} & \textbf{0.657} \\
\midrule

\multicolumn{2}{c|}{\textbf{Avg (S + SF)}} 
& \textbf{22.834} & \textbf{0.665} & \textbf{0.222} & \textbf{68.343} & \textbf{0.384} & \textbf{0.666} \\

\bottomrule
\end{tabular}
}
\vspace{-8pt}
\caption{Image quality assessment results across different prompt designs. We report group-wise averages (bold) and combined averages for long (L+LF) and short (S+SF) prompts.}
\label{tab:iqa}
\vspace{-18pt}
\end{table}

\begin{figure}[tp]
    \centering
    \includegraphics[width=\linewidth]{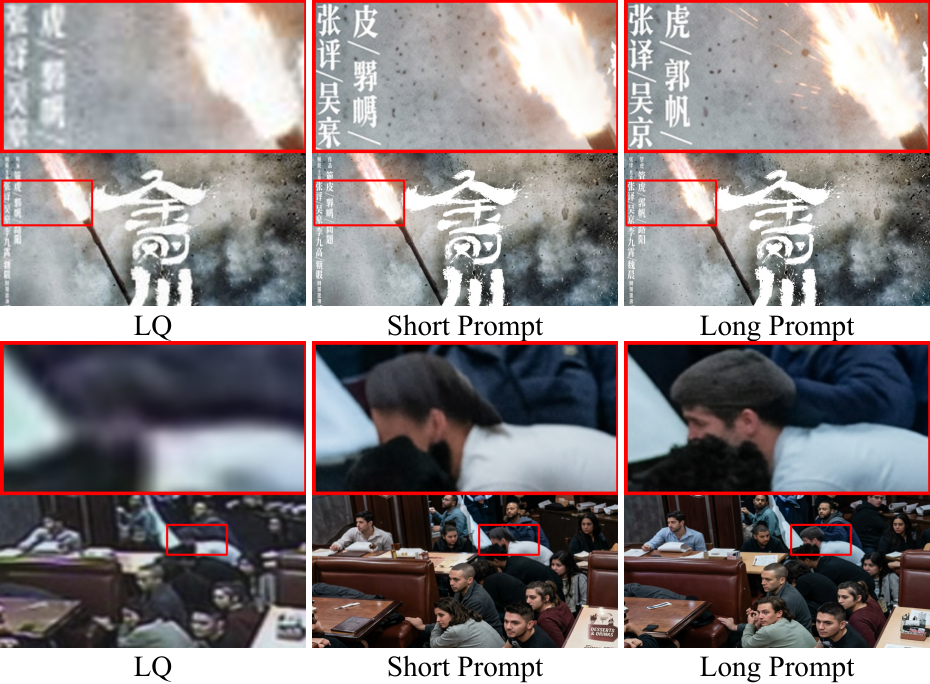}
    \vspace{-18pt}
    \caption{Qualitative comparison of prompt length. Long prompts yield more accurate and detailed restorations than short prompts, especially for challenging scenarios such as text and surveillance.}
    \label{fig3}
    \vspace{-21pt}
\end{figure}

\vspace{-8pt}
\section{Experimental Setup}
\vspace{-4pt}
\noindent\textbf{Dataset.} To systematically evaluate the IR capability of Nano Banana 2, experiments are conducted on a dataset derived from a prior work~\cite{yin2026far}, which provides a diverse collection of scene categories and image degradation types. 
From this dataset, a subset is selected to cover challenging visual structures and realistic degradation scenarios in practical restoration tasks.
Specifically, 13 representative scene categories are included: Aerial View, Animal Fur, Architecture, Cartoon/Comic, Large Face, Medium Face, Small Face, Crowd, Text, Hands/Feet, Trees/Leaves, Hand-drawn, and Fabric Texture. 
These categories encompass both natural and structured content, involving fine-grained details such as hair, text, dense textures, and small objects, which are particularly challenging for accurate reconstruction.
In terms of degradation, seven representative types are considered: Defocus Blur, Motion Blur, Digital Zoom, Old Film, Old Photo (Black and White), Old Photo (Color), and Surveillance.
These degradations span both optical distortions and quality degradation commonly observed in historical imagery, low-quality imaging systems, and real-world compressed data. 
Such a diverse selection enables a comprehensive evaluation of restoration performance across varied degradation characteristics.
All images are of a resolution of 1024$\times$1024 for evaluation to ensure consistent input conditions across different models.

\noindent\textbf{Prompt Design.} 
Effective prompt design is crucial for guiding Nano Banana 2 to perform high-quality image restoration.
To study the effect of prompt formulation, we design 12 prompts with different structures and levels of detail, as shown in~\cref{tab:prompts}.
The prompts are divided into two length groups: short prompts containing 26--30 words, and long prompts containing 34--47 words. 
Since a successful restoration should be both perceptually plausible and faithful to the original image content, we further vary whether fidelity preservation is explicitly emphasized within each length group.
As a result, we differentiate prompts in each length group based on whether they explicitly emphasize fidelity preservation, resulting in prompts that either include or omit fidelity-related descriptions.
This design enables us to analyze how prompt length and fidelity-oriented instructions influence restoration performance. 
In addition, to evaluate the stability of the generative model, we select four representative prompts and perform three repeated generations for each input image under identical conditions. 
This experiment measures the consistency of outputs produced by the model.
Finally, the prompt that achieves the best overall performance in preliminary experiments is applied to the entire dataset to evaluate Nano Banana 2 under different scenes and degradation types.

\noindent\textbf{Compared Methods.} 
We compare Nano Banana 2 with several SOTA IR approaches, including HYPIR~\cite{lin2025harnessing}, TSD-SR~\cite{tong2023tsdsr}, PiSA-SR~\cite{sun2025pixel} and DiffBIR~\cite{lin2024diffbir}. 
These methods leverage powerful generative priors to reconstruct realistic image details and have demonstrated strong performance on multiple restoration tasks.

\noindent\textbf{Evaluation Metrics.} 
To evaluate restoration performance comprehensively, we adopt both full-reference (FR) and no-reference (NR) image quality assessment (IQA) metrics.
For FR evaluation, we report PSNR, SSIM~\cite{wang2004image}, and LPIPS~\cite{zhang2018unreasonable}, which measure pixel-level fidelity and structural consistency between restored images and their corresponding ground-truth references. These metrics are standard in IR benchmarks and provide a reliable assessment of reconstruction accuracy.
To further assess perceptual quality without relying on ground-truth images, we employ NR metrics including MUSIQ~\cite{ke2021musiq}, MANIQA~\cite{yang2022maniqa}, and CLIP-IQA~\cite{wang2023exploring}. These metrics are designed to better align with human visual perception and are particularly suitable for evaluating generative restoration models that emphasize perceptual realism over strict pixel-wise consistency.
By combining FR and NR metrics, our evaluation captures complementary aspects of restoration performance, including distortion fidelity, structural consistency, and perceptual quality, enabling a more comprehensive comparison between Nano Banana 2 and existing restoration methods.

\begin{figure}[tp]
    \centering
    \includegraphics[width=\linewidth]{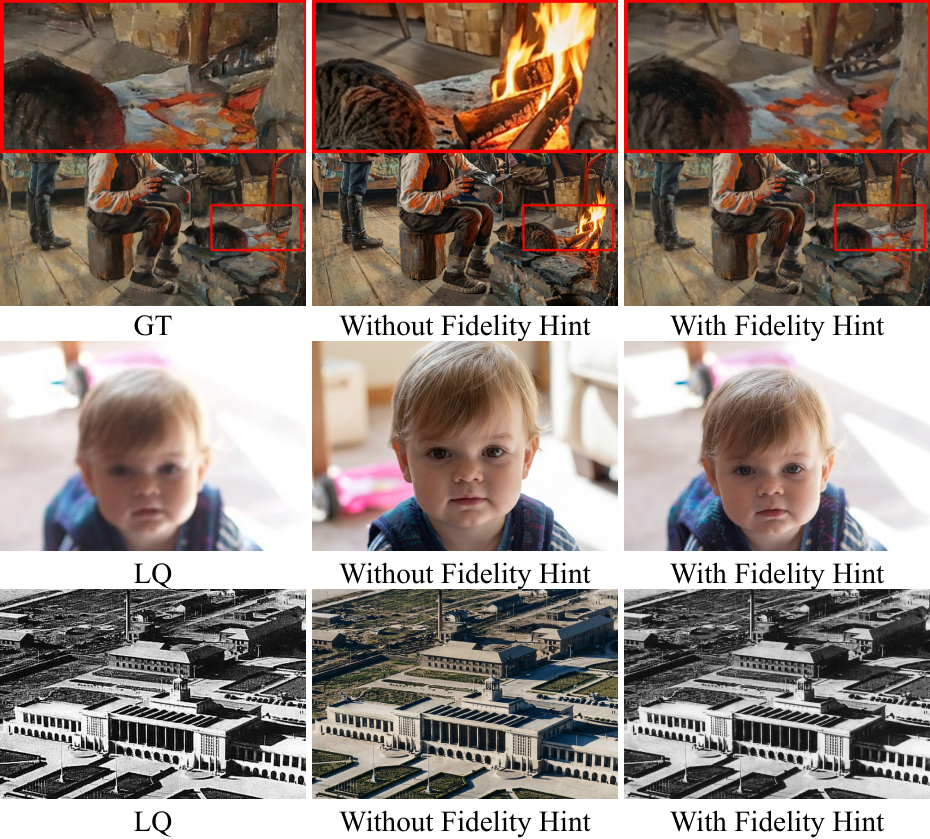}
    \vspace{-20pt}
    \caption{Effect of fidelity constraints. Prompts without fidelity introduce semantic artifacts, while fidelity-constrained prompts preserve structure and semantics.}
    \label{fig4}
    \vspace{-22pt}
\end{figure}

\begin{figure}[tp]
    \centering
    \includegraphics[width=\linewidth]{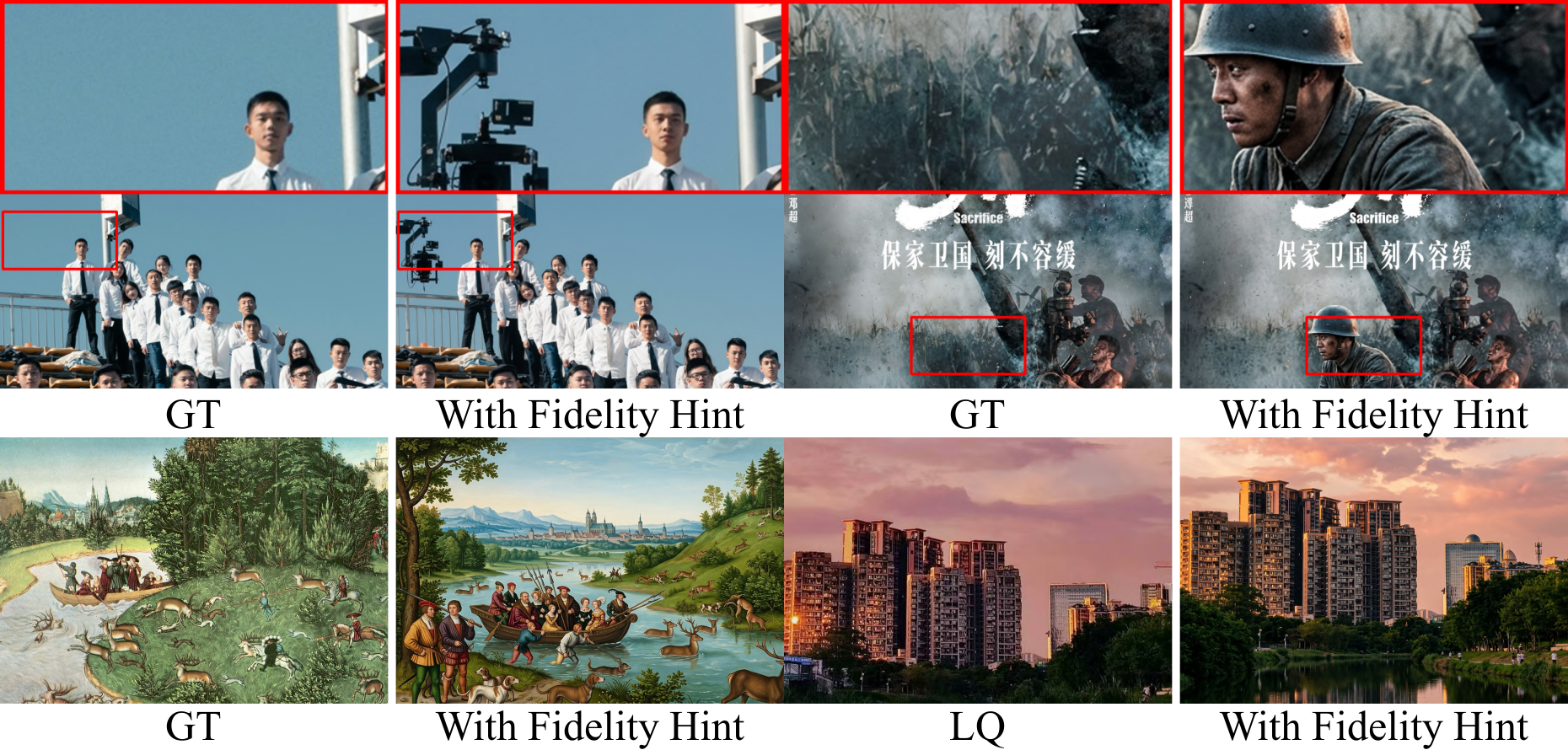}
    \vspace{-18pt}
    \caption{Failure cases with fidelity constraints. Infidelity still occurs despite explicit fidelity guidance.}
    \label{fig6}
    \vspace{-18pt}
\end{figure}

\vspace{-8pt}
\section{Results and Analysis}
\vspace{-4pt}
\subsection{Impact of Prompt Design}
\vspace{-3pt}
\label{sec:prompt impact}
To investigate the effect of different prompts on Nano Banana 2 for image restoration, we evaluate 12 prompts with different lengths and fidelity constraints. Specifically, prompts are divided into long prompts (L) and short prompts (S), and further categorized based on whether they explicitly include fidelity-related hints (F).

\noindent\textbf{Effect of Prompt Length.}
We first analyze the impact of prompt length by comparing the average performance across all prompt variants. Specifically, we aggregate results over six long prompts (three with fidelity constraints and three without) and six short prompts with the same configuration.
As shown in~\cref{tab:iqa}, the average score of long prompts consistently outperforms short prompts across all six IQA metrics. In the FR setting, long prompts achieve higher PSNR and SSIM and lower LPIPS, indicating improved reconstruction fidelity and structural consistency. Similarly, in the NR setting, long prompts yield better MUSIQ, MANIQA, and CLIP-IQA scores, suggesting enhanced perceptual quality.
These results indicate that increasing prompt length provides more comprehensive guidance for the restoration process, leading to consistent improvements in both distortion-based and perceptual metrics.
Qualitative comparisons further support these findings. As illustrated in~\cref{fig3}, long prompts produce more accurate and detailed restorations in challenging scenarios such as text and surveillance imagery. For instance, in text restoration, long prompts correctly recover actor names with clear and consistent characters, whereas short prompts exhibit noticeable errors. In surveillance scenes, long prompts recover finer facial details with improved clarity, while short prompts tend to produce overly smooth or blurred results.

\noindent\textbf{Effect of Fidelity Instructions.}
We further examine the role of fidelity-related descriptions in prompts. When fidelity constraints are introduced, restoration performance improves noticeably for short prompts. In particular, the SF group (short prompts with fidelity instructions) achieves the best FR results among all prompt types, suggesting that explicit instructions emphasizing detail preservation and structural consistency can effectively guide the model toward more faithful reconstruction.
To further analyze this effect, we conduct a case-level evaluation of infidelity, which we define as severe semantic deviations from the input, including object insertion, significant shape distortion, or inconsistent image semantics. Across 12 prompt variants and 35 test images, prompts without fidelity constraints produce on average 2 infidelity cases per prompt, while this number is reduced to 0.5 cases when fidelity instructions are introduced. This result provides direct evidence that fidelity-related prompts effectively suppress undesired semantic alterations and improve structural consistency.
Qualitative examples further illustrate this phenomenon. As shown in~\cref{fig4}, prompts without fidelity constraints often introduce noticeable semantic artifacts, including spurious object generation (e.g., artificial flame-like structures), unintended semantic shifts, and incorrect colorization (e.g., transforming grayscale inputs into colorized outputs). In contrast, when fidelity constraints are applied, the restored results better preserve the original scene structure and semantic content, avoiding such distortions.
However, we also observe that infidelity is not completely eliminated even with fidelity constraints in~\cref{fig6}. Occasional failure cases still occur, indicating that prompt-based guidance alone is insufficient to fully constrain the generative process.
In contrast, prompts without fidelity constraints tend to produce outputs that deviate more from the ground truth. For instance, the prompt S3 obtains the highest scores across all no-reference IQA metrics, while simultaneously achieving the lowest scores in full-reference metrics. This observation indicates that the model tends to generate visually pleasing textures and enhanced details when fidelity constraints are absent. Although such outputs appear perceptually appealing, they introduce deviations from the ground-truth structure, leading to lower PSNR and SSIM.

\begin{figure}[tp]
    \centering
    \includegraphics[width=\linewidth]{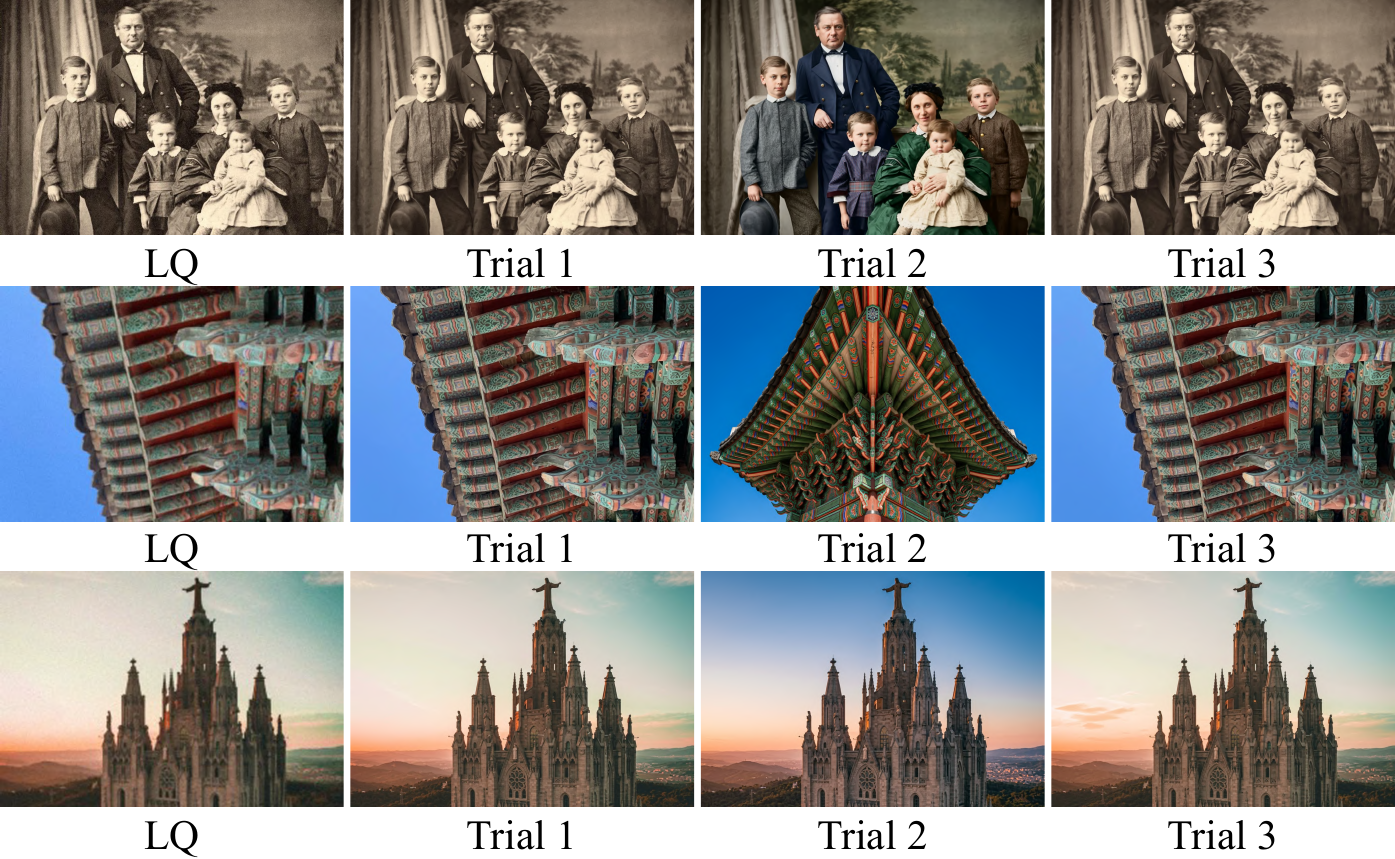}
    \vspace{-20pt}
    \caption{Failure cases in output stability. Repeated runs on the same input produce inconsistent results, including color shifts, scale variations, and structural changes.}
    \label{fig5}
    \vspace{-16pt}
\end{figure}

\begin{table}[tp]
\centering
\scriptsize
\setlength{\tabcolsep}{3.5pt}
\renewcommand{\arraystretch}{1.0}

\resizebox{\columnwidth}{!}{
\begin{tabular}{c|c|cccccc}
\toprule
Prompt & Type 
& PSNR & SSIM & LPIPS & MUSIQ & MANIQA & CLIP-IQA \\
\midrule

\multirow{2}{*}{L3}
& Stat & 1.238 & 0.857 & 0.857 & 3.257 & 0.171 & 0.229 \\
& p-val & 0.538 & 0.651 & 0.651 & 0.196 & 0.918 & 0.892 \\
\midrule

\multirow{2}{*}{LF1}
& Stat & 1.238 & 1.524 & 0.095 & 6.229 & 3.486 & 1.429 \\
& p-val & 0.538 & 0.467 & 0.953 & \textbf{0.044} & 0.175 & 0.490 \\
\midrule

\multirow{2}{*}{S1}
& Stat & 0.667 & 1.238 & 2.000 & 2.114 & 4.800 & 7.943 \\
& p-val & 0.717 & 0.538 & 0.368 & 0.347 & 0.091 & \textbf{0.019} \\
\midrule

\multirow{2}{*}{SF2}
& Stat & 0.667 & 2.667 & 1.238 & 3.257 & 0.057 & 0.914 \\
& p-val & 0.717 & 0.264 & 0.538 & 0.196 & 0.972 & 0.633 \\

\bottomrule
\end{tabular}
}
\vspace{-8pt}
\caption{Friedman test results across prompt designs. Stat and p-val denote the Friedman statistic and corresponding p-value. Statistically significant results ($p < 0.05$) are highlighted in bold.}
\label{tab:significance}
\vspace{-20pt}
\end{table}

\noindent\textbf{Output Stability.}
We further evaluate the stability of Nano Banana 2 under repeated inference. For each prompt type, we select representative prompts, including L3, LF1, S1, and SF2, and perform three independent runs for each input image under identical conditions. For each run, we compute six IQA metrics and conduct a \textit{Friedman test}~\cite{friedman1937use} to examine whether the differences across repeated outputs are statistically significant. The null hypothesis assumes that there is no significant difference among the repeated runs for the same input and prompt.
Overall, \cref{tab:significance} indicates that the model exhibits stable behavior in most cases. Across the four prompt groups and six IQA metrics, we obtain a total of 24 statistical tests, among which only 2 cases show statistical significance ($p < 0.05$), leading to the rejection of the null hypothesis. In contrast, the remaining majority have $p$-values far greater than 0.05, indicating that the null hypothesis cannot be rejected. This suggests that the variations across repeated runs are generally insignificant, and the model produces consistent restoration results despite the stochastic nature of the generative process. This observation holds for both full-reference and no-reference metrics, indicating stability in terms of both reconstruction fidelity and perceptual quality.
However, we also observe occasional outliers where repeated runs lead to large variations in IQA scores. These cases typically occur for challenging inputs with complex structures or severe degradations, where the generative process exhibits increased stochasticity. As shown in~\cref{fig5}, repeated trials on the same input can produce noticeably different outputs, including shifts in dominant object color and structural changes, often accompanied by inconsistent texture synthesis and metric fluctuations.
These results indicate that, although Nano Banana 2 demonstrates overall stability under repeated inference, its outputs remain sensitive to stochastic sampling in difficult scenarios. This suggests that controlling generative randomness is crucial for improving reliability, highlighting stability as an important direction for future research in generative image restoration.

\begin{table*}[tp]
\centering
\scriptsize
\setlength{\tabcolsep}{6.5pt}
\begin{tabular}{ccccccccccccc}
\toprule
\multirow{2}{*}{Method} & \multicolumn{6}{c}{Small Faces} & \multicolumn{6}{c}{Hands/Feet} \\
\cmidrule(lr){2-7} \cmidrule(lr){8-13} 
& PSNR & SSIM & LPIPS$\downarrow$ & MUSIQ & MANIQA & CLIP-IQA & PSNR & SSIM & LPIPS$\downarrow$ & MUSIQ & MANIQA & CLIP-IQA \\
\midrule
Nano Banana 2& \textbf{23.809} & \textbf{0.735} & \textbf{0.146} & 72.989 & 0.427 & 0.662 
& \textbf{26.456} & \textbf{0.763} & \textbf{0.148} & 70.032 & 0.431 & 0.683 \\
HYPIR& 20.857 & 0.666 & 0.176 & 73.781 & 0.481 & 0.550 
& 24.280 & 0.708 & \underline{0.178} & 71.728 & 0.501 & 0.634 \\
PiSA-SR& \underline{22.518} & \underline{0.676} & \underline{0.175} & 75.458 & 0.545 & 0.705 
& \underline{25.828} & \underline{0.749} & 0.184 & \underline{75.013} & \underline{0.555} & 0.732 \\
TSD-SR& 21.375 & 0.654 & 0.183 & \textbf{76.590} & \textbf{0.593} & \textbf{0.737} 
& 25.482 & 0.734 & 0.186 & 74.390 & 0.535 & \underline{0.745} \\
DiffBIR& 21.580 & 0.662 & 0.203 & \underline{75.725} & \underline{0.559} & \underline{0.710} 
& 25.362 & 0.732 & 0.204 & \textbf{75.507} & \textbf{0.594} & \textbf{0.749} \\
\midrule

\multirow{2}{*}{Method} & \multicolumn{6}{c}{Text} & \multicolumn{6}{c}{Motion Blur} \\
\cmidrule(lr){2-7} \cmidrule(lr){8-13} 
& PSNR & SSIM & LPIPS$\downarrow$ & MUSIQ & MANIQA & CLIP-IQA & PSNR & SSIM & LPIPS$\downarrow$ & MUSIQ & MANIQA & CLIP-IQA \\
\midrule
Nano Banana 2& \underline{20.347} & \textbf{0.659} & 0.230 & 69.136 & 0.455 & \textbf{0.716} 
& / & / & / & 55.313 & 0.278 & 0.517 \\
HYPIR& 20.007 & \underline{0.643} & \underline{0.228} & 66.680 & 0.443 & 0.601 
& / & / & / & 47.633 & 0.284 & 0.444 \\
PiSA-SR& \textbf{21.551} & 0.642 & 0.247 & \underline{70.474} & 0.482 & 0.597 
& / & / & / & 54.016 & \underline{0.332} & 0.514 \\
TSD-SR& 20.167 & 0.616 & \textbf{0.227} & \textbf{71.338} & \textbf{0.527} & 0.659 
& / & / & / & \textbf{56.043} & 0.315 & \textbf{0.529} \\
DiffBIR& 19.844 & 0.609 & 0.271 & 69.311 & \underline{0.496} & \underline{0.690} 
& / & / & / & \underline{55.495} & \textbf{0.353} & \underline{0.525} \\
\midrule

\multirow{2}{*}{Method} & \multicolumn{6}{c}{Old Film} & \multicolumn{6}{c}{Surveillance} \\
\cmidrule(lr){2-7} \cmidrule(lr){8-13} 
& PSNR & SSIM & LPIPS$\downarrow$ & MUSIQ & MANIQA & CLIP-IQA & PSNR & SSIM & LPIPS$\downarrow$ & MUSIQ & MANIQA & CLIP-IQA \\
\midrule
Nano Banana 2& / & / & / & 70.682 & 0.388 & \underline{0.710} 
& / & / & / & 65.514 & 0.346 & 0.555 \\
HYPIR& / & / & / & 66.447 & 0.360 & 0.522 
& / & / & / & 55.320 & 0.287 & 0.384 \\
PiSA-SR& / & / & / & \underline{70.823} & 0.423 & 0.686 
& / & / & / & \textbf{67.773} & \textbf{0.405} & \underline{0.577} \\
TSD-SR& / & / & / & \textbf{71.979} & \underline{0.437} & \textbf{0.725} 
& / & / & / & \underline{66.654} & 0.389 & \textbf{0.617} \\
DiffBIR& / & / & / & 69.159 & \textbf{0.440} & 0.640 
& / & / & / & 61.437 & \underline{0.396} & 0.547 \\
\bottomrule
\end{tabular}
\vspace{-8pt}
\caption{Quantitative comparison across diverse challenging scenarios and degradations, including Small Faces, Hands/Feet, Text, Motion Blur, Old Film, and Surveillance. Best results are in \textbf{bold}, and second-best results are \underline{underlined}.}
\label{tab:1}
\vspace{-16pt}
\end{table*}

\noindent\textbf{Discussion.}
The above observations reveal a clear perception–distortion trade-off in prompt-guided restoration. Prompts without fidelity constraints encourage the model to prioritize perceptual enhancement, which improves no-reference quality metrics but reduces fidelity to the original image. In contrast, prompts emphasizing fidelity constrain the generative process and lead to more accurate reconstruction, improving distortion-based metrics.
Overall, our results suggest that concise prompts with explicit fidelity instructions provide the most effective guidance for image restoration. This finding highlights the importance of prompt design in generative restoration systems and suggests that prompt can significantly influence the balance between perceptual quality and reconstruction fidelity.

\vspace{-4pt}
\subsection{Performance Across Scenes and Degradations}
\vspace{-3pt}
To evaluate the generalization of Nano Banana 2, we conduct experiments across diverse scenes and degradations using a unified prompt setting (LF3). The evaluation covers multiple scene categories and degradation conditions, and compare its performance with several SOTA methods.

We analyze model performance on several challenging scenarios, including Small Faces, Crowd, Hands/Feet, and Text, as well as degradations such as Motion Blur, Old Film, and Surveillance, as shown in~\cref{tab:1}. As illustrated in~\cref{fig2}, Nano Banana 2 consistently produces visually appealing results with enhanced sharpness and rich details. It reconstructs plausible global structures under severe degradations and yields perceptually cleaner outputs in low-quality settings such as surveillance images. The model also generates recognizable facial patterns, plausible geometry for hands and feet, and improved text legibility.

However, a closer inspection reveals a clear gap between perceptual quality and restoration fidelity. Nano Banana 2 tends to over-enhance details, leading to over-generation, hallucinated textures, color shifts, and background inconsistencies. While visually sharper, these results do not always faithfully preserve low-level details or remain consistent with the input. In contrast, IR models produce more conservative outputs that better maintain input-aligned structures and color distributions, albeit with less perceptual sharpness. To further illustrate these differences, we provide annotated visual comparisons and analysis examples on our project website.

\begin{table}[tp]
\centering
\small
\setlength{\tabcolsep}{4pt}
\renewcommand{\arraystretch}{0.95}

\resizebox{\columnwidth}{!}{
\begin{tabular}{c|cccccc}
\toprule
Method & PSNR & SSIM & LPIPS$\downarrow$ & MUSIQ & MANIQA & CLIP-IQA \\
\midrule

HYPIR  & 21.307 & 0.622 & 0.240 & 67.103 & 0.407 & 0.582 \\
PiSA-SR & \textbf{22.744} & \underline{0.633} & 0.237 & \underline{71.276} & 0.468 & 0.681 \\
TSD-SR  & 21.439 & 0.599 & \underline{0.232} & \textbf{72.144} & \textbf{0.479} & \textbf{0.708} \\
DiffBIR& 21.856 & 0.602 & 0.273 & 69.974 & \underline{0.478} & \underline{0.685} \\
Nano Banana 2 & \underline{22.541} & \textbf{0.649} & \textbf{0.222} & 68.841 & 0.394 & 0.676 \\

\bottomrule
\end{tabular}
}
\vspace{-8pt}
\caption{Comparison with existing methods on IR benchmarks. Nano Banana 2 achieves competitive distortion metrics while maintaining strong perceptual quality. The best and second best results are \textbf{bold} and \underline{underlined}, respectively.}
\label{tab:comparison}
\vspace{-15pt}
\end{table}

\vspace{-4pt}
\subsection{Comparison with State-of-the-Art Methods}
\vspace{-3pt}
We compare Nano Banana 2 with several representative IR models, including recent generative and learning-based approaches, such as HYPIR, PISA-SR, TSD-SR, and DiffBIR. These methods cover different restoration paradigms, providing a comprehensive benchmark for evaluating the performance of Nano Banana 2.

\noindent\textbf{Quantitative Comparison.}
We first report quantitative results using both FR and NR IQA metrics, as shown in~\cref{tab:comparison}. Nano Banana 2 achieves competitive performance on these metrics, particularly obtaining favorable SSIM and LPIPS scores compared to other methods. 
However, these results should be interpreted with caution. While FR metrics suggest strong structural similarity to ground truth, they are not sufficiently sensitive to hallucinated details, local inconsistencies, or color deviations introduced during generation~\cite{jinjin2020pipal,hu2026position,yu2024scaling,lin2025harnessing,yin2026far}. In practice, we observe that Nano Banana 2 often enhances textures and sharpness in a way that improves metric scores, yet does not necessarily correspond to faithful recovery of the underlying signal. In contrast, traditional IR methods tend to produce more conservative outputs that better preserve input-consistent low-level details, despite sometimes receiving lower metric scores.

\begin{figure}[tp]
    \centering
    \includegraphics[width=\linewidth]{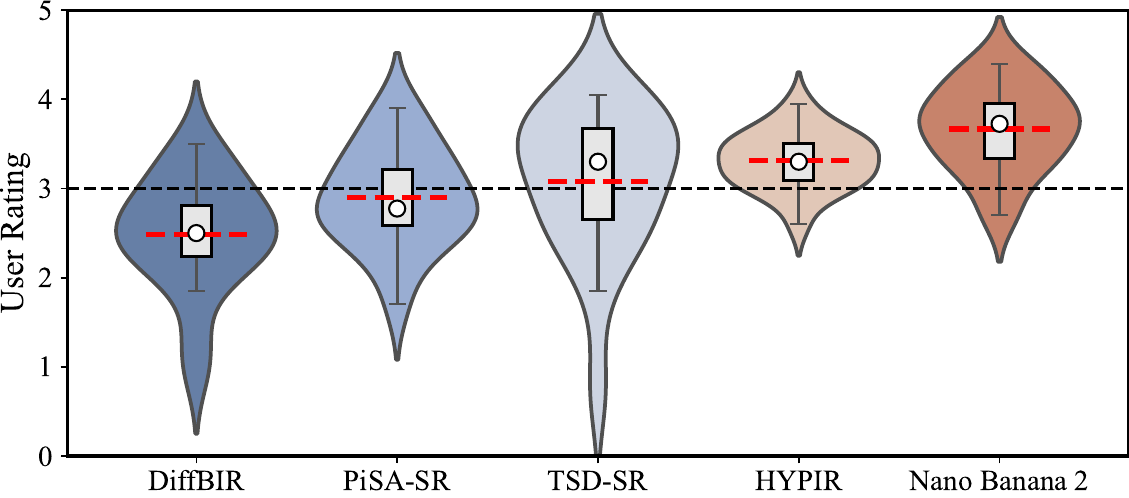}
    \vspace{-18pt}
    \caption{User study results. Nano Banana 2 achieves the highest average score and shows more consistent perceptual quality.}
    \label{fig1}
    \vspace{-14pt}
\end{figure}

\begin{figure}[tp]
    \centering
    \includegraphics[width=\linewidth]{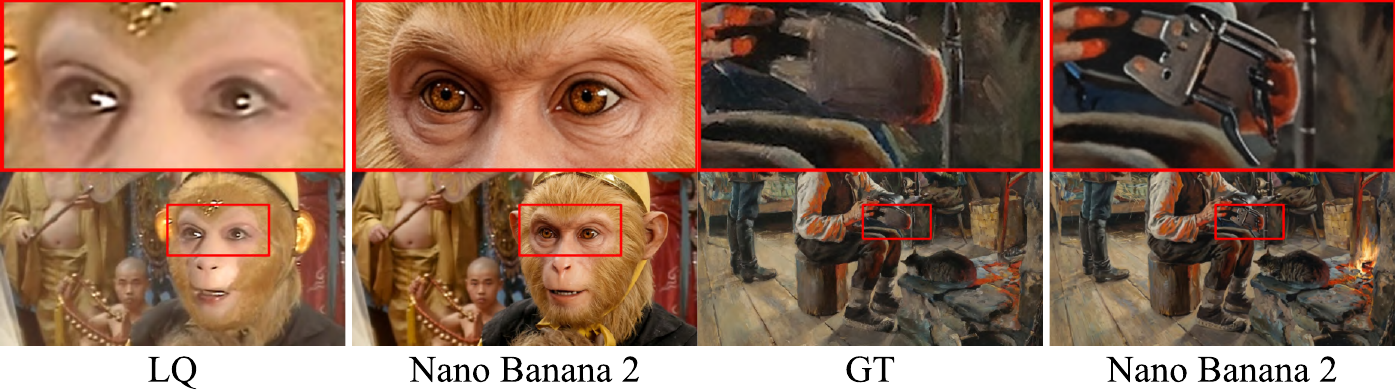}
    \vspace{-18pt}
    \caption{Over-generation artifacts. The model produces excessive details such as amplified textures and unrealistic patterns.}
    \label{fig7}
    \vspace{-18pt}
\end{figure}

\begin{figure*}[tp] 
    \centering
    \includegraphics[width=\textwidth]{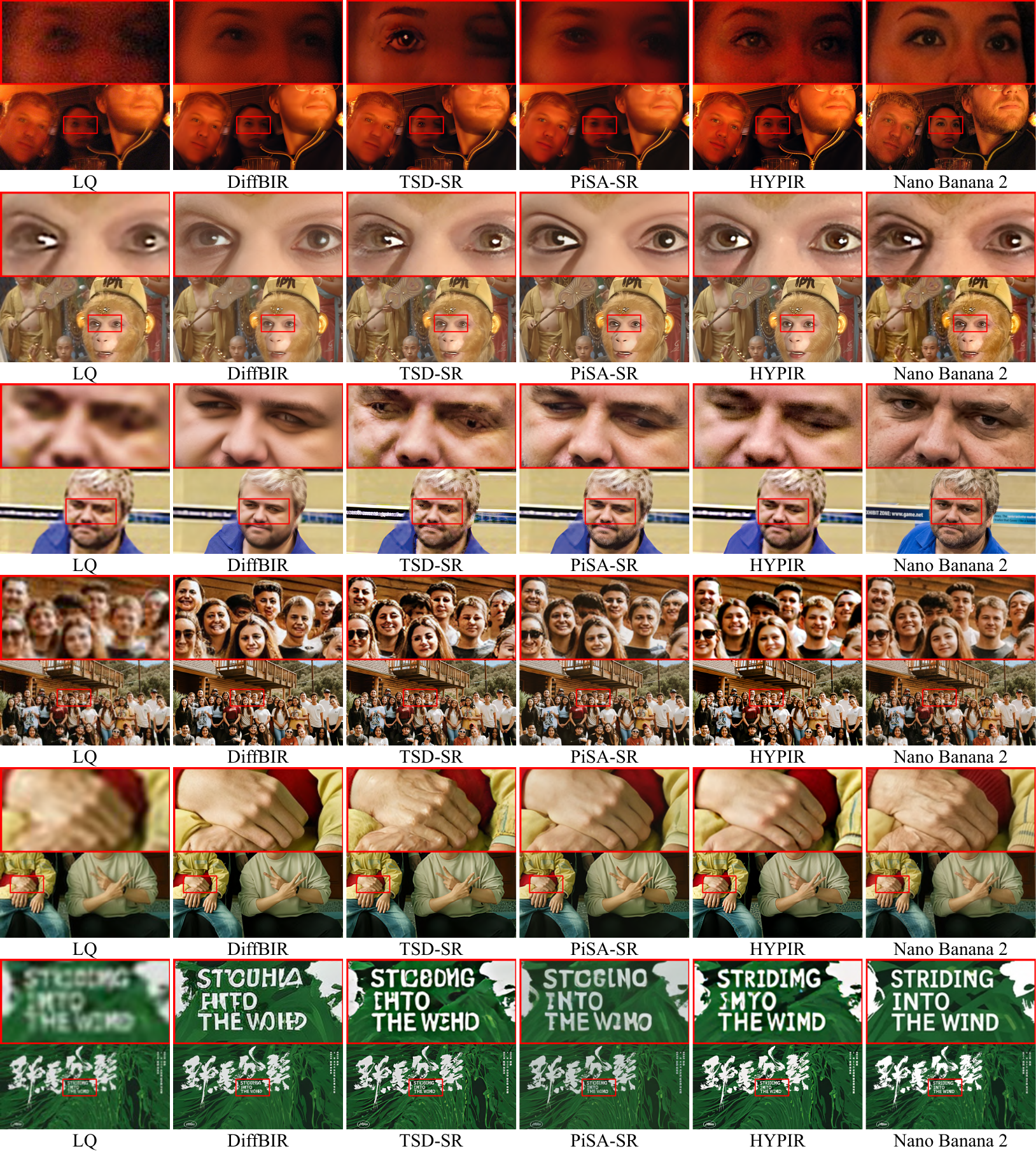} 
    \vspace{-20pt}
    \caption{Qualitative comparison on challenging scenarios and degradations. Nano Banana 2 produces clearer and more detailed results across motion blur, old film, surveillance, small face, hands and feet, and text, but may introduce over-enhanced details and inconsistencies.}
    \label{fig2}
    \vspace{-18pt}
\end{figure*}

\noindent\textbf{User Study.}
Given the limitations of IQA metrics in assessing perceptual quality~\cite{jinjin2020pipal, yu2024scaling, hu2026position}, we further conduct a user study from a human perspective. We recruit 20 participants to evaluate restoration results produced by five models, including Nano Banana 2 and four baselines, on 20 images spanning diverse scenes and degradation types. Participants are asked to rate each result on a scale from 0 (worst) to 5 (best) based on overall visual quality.
As summarized in~\cref{fig1}, Nano Banana 2 achieves the highest average score and exhibits a relatively concentrated score distribution, indicating stable perceptual performance across samples. This suggests that its outputs are generally preferred by human observers. 
Nevertheless, a closer examination reveals that user preference is strongly biased toward visually rich and sharp results, often favoring images with enhanced or synthesized details. As a result, the user study primarily reflects perceptual attractiveness rather than restoration fidelity. In particular, participants rarely penalize inconsistencies with the degraded input, such as color shifts, background alterations, or over-generated textures.

\vspace{-9pt}
\section{Limitation and Discussion}
\vspace{-4pt}
\noindent\textbf{Failure Cases.}
Despite its strong overall performance, Nano Banana 2 still exhibits limitations under challenging conditions. We categorize the failure cases into two main types: infidelity and over-generation.
Infidelity refers to severe semantic inconsistencies between the restored image and the input, as illustrated in~\cref{fig6}. In complex scenes, Nano Banana 2 may introduce non-existent objects or persons, alter the primary structure, or significantly deviate from the original semantic content. Such issues are particularly common in scenarios involving small faces, dense crowds, or heavily degraded inputs, where reliable structural recovery is inherently difficult.
Over-generation, on the other hand, describes the tendency of the model to produce excessive or exaggerated details, as shown in~\cref{fig7}. This includes overly dense hair, amplified textures, or unrealistic fine-grained patterns, which may lead to visually inconsistent or implausible results. While these outputs may appear sharp at first glance, they often reduce overall coherence and realism.
These failure modes highlight the challenges of controlling generative restoration models, especially in complex scenarios where both semantic fidelity and perceptual plausibility are critical.

\noindent\textbf{Prompt Sensitivity and Engineering.}
Our experiments reveal that the restoration performance of Nano Banana 2 is highly sensitive to prompt design. As demonstrated in~\cref{sec:prompt impact}, different prompt formulations can lead to significantly different behaviors, ranging from faithful reconstruction to perceptual enhancement with hallucinated details. In particular, prompts without fidelity constraints tend to produce visually appealing but structurally inconsistent results, while prompts emphasizing fidelity improve reconstruction accuracy but may reduce perceptual sharpness.
To achieve optimal performance, careful prompt engineering is often required. In practice, concise prompts with explicit fidelity-related instructions generally provide the most reliable results. However, even with well-designed prompts, the model may still exhibit variability across runs, especially in complex scenarios. In some cases, iterative refinement or multi-round prompting is necessary to obtain satisfactory outputs, indicating that restoration with generative models is not yet a fully deterministic process.

\noindent\textbf{Can General-Purpose Models Serve as Unified IR Solvers?}
A central question raised in this work is whether a general-purpose image editing model can serve as a unified solver for image restoration. Based on our findings, the answer is \emph{conditionally affirmative}. Nano Banana 2 demonstrates strong adaptability across diverse scenes and degradation types, producing visually plausible results under a unified prompt setting. This suggests that general-purpose generative models possess the capability to handle a wide range of restoration tasks within a single framework, especially when structural recovery and perceptual quality are the primary objectives. However, this capability should be interpreted with caution. Our analysis reveals that Nano Banana 2 exhibits a clear tendency toward perceptual enhancement rather than faithful restoration, often introducing over-generated details, color deviations, and background inconsistencies. While such outputs are frequently preferred in user studies and can achieve competitive scores on existing IQA metrics, they do not necessarily reflect accurate recovery of the degraded input. This highlights a fundamental misalignment between current evaluation protocols and restoration fidelity.

\vspace{-10pt}
\section{Conclusion}
\vspace{-5pt}
In this work, we present a systematic evaluation of Nano Banana 2 for IR across diverse scenes and degradation types. 
Our results show that prompt design plays a critical role in controlling restoration behavior. 
However, we observe a consistent gap between perceptual quality and restoration fidelity: Nano Banana 2 tends to produce visually rich results while sacrificing input consistency, especially under severe degradations. This discrepancy is not well captured by existing IQA metrics or user studies, which are biased toward perceptual preference. 
Overall, while general-purpose generative models show promise as unified IR solvers, their strengths currently lie in perceptual enhancement rather than faithful reconstruction, highlighting the need for improved controllability and evaluation protocols.

\newpage
\noindent\textbf{Acknowledgment.} This work was supported by the National Natural Science Foundation of China (Grant No. 62276251).

{
    \small
    \bibliographystyle{unsrt}
    \bibliography{main}
}

\maketitlesupplementary
\appendix
\section*{Appendix}
While Nano Banana 2 demonstrates strong structural reconstruction ability and improves over earlier generative models, it still falls short of traditional IR methods in preserving fine-grained low-level details. Importantly, such discrepancies are not adequately captured by existing IQA metrics or user studies. This appendix provides deeper empirical evidence and analysis to support the key observation in the main paper: the inherent gap between perceptual quality and restoration fidelity in generative IR models.

\section{Visual Analysis}

\subsection{The Limitation of Metrics and User Study}
\begin{figure*}[tp] 
    \centering
    \includegraphics[width=\textwidth]{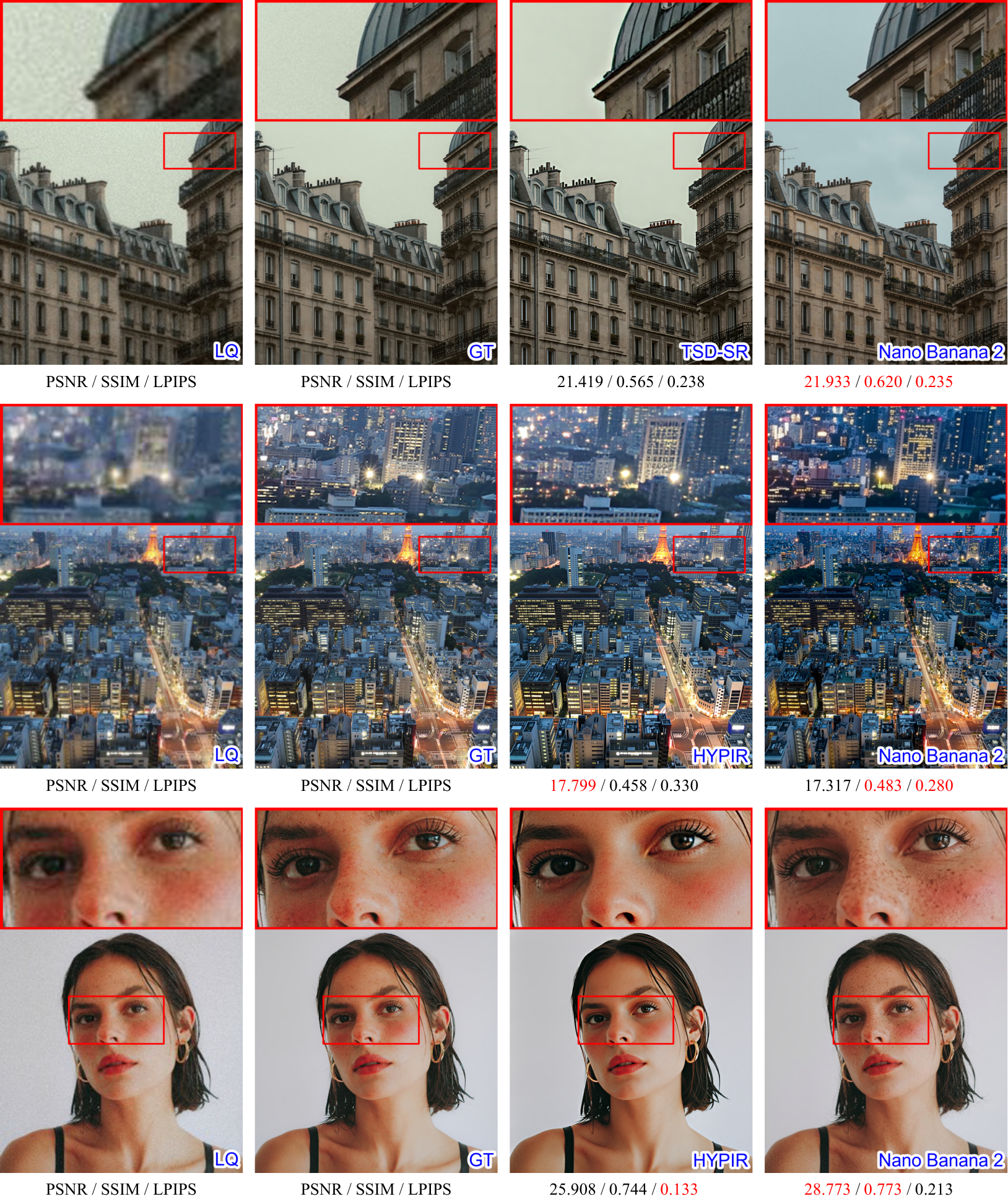} 
    \vspace{-15pt}
    \caption{FR IQA metrics such as PSNR, SSIM, and LPIPS often fail to reflect perceptual restoration quality. Nano Banana 2 may still exhibit noticeable color shifts, tonal inconsistencies, or hallucinated details despite achieving favorable metric scores. Higher values indicate better performance for PSNR and SSIM, while lower values are preferred for LPIPS. The best result for each metric is highlighted in red. Zoom in for better observation.}
    \label{fr_iqa}
    \vspace{-18pt}
\end{figure*}

Although Nano Banana 2 achieves competitive scores in IQA metrics, these metrics are insufficient to capture subtle yet critical discrepancies in low-level fidelity~\cite{yu2024scaling, hu2026position}. 
In several examples~\cref{fr_iqa}, results of Nano Banana 2 achieve higher PSNR or SSIM, or lower LPIPS, but produce visually inferior results.
Specifically, these results exhibit noticeable color shifts and tonal inconsistencies compared to the ground truth, despite receiving competitive metric scores. This behavior is consistent with prior studies~\cite{lee2015towards} showing that conventional FR metrics are primarily dominated by structural and luminance fidelity, while exhibiting limited sensitivity to chrominance distortions and perceptual color consistency.
In other cases, methods introduce hallucinated textures or over-sharpened details that appear visually unnatural but remain weakly penalized by existing FR metrics. Such discrepancies are related to the well-known perception-distortion tradeoff~\cite{blau2018perception}, where distortion-oriented metrics may favor structurally similar reconstructions even when perceptual realism is degraded.
These observations suggest that high FR IQA scores do not necessarily correspond to faithful perceptual restoration, particularly when structural consistency is preserved but low-level appearance statistics deviate from human visual preference.

\begin{figure*}[tp] 
    \centering
    \includegraphics[width=\textwidth]{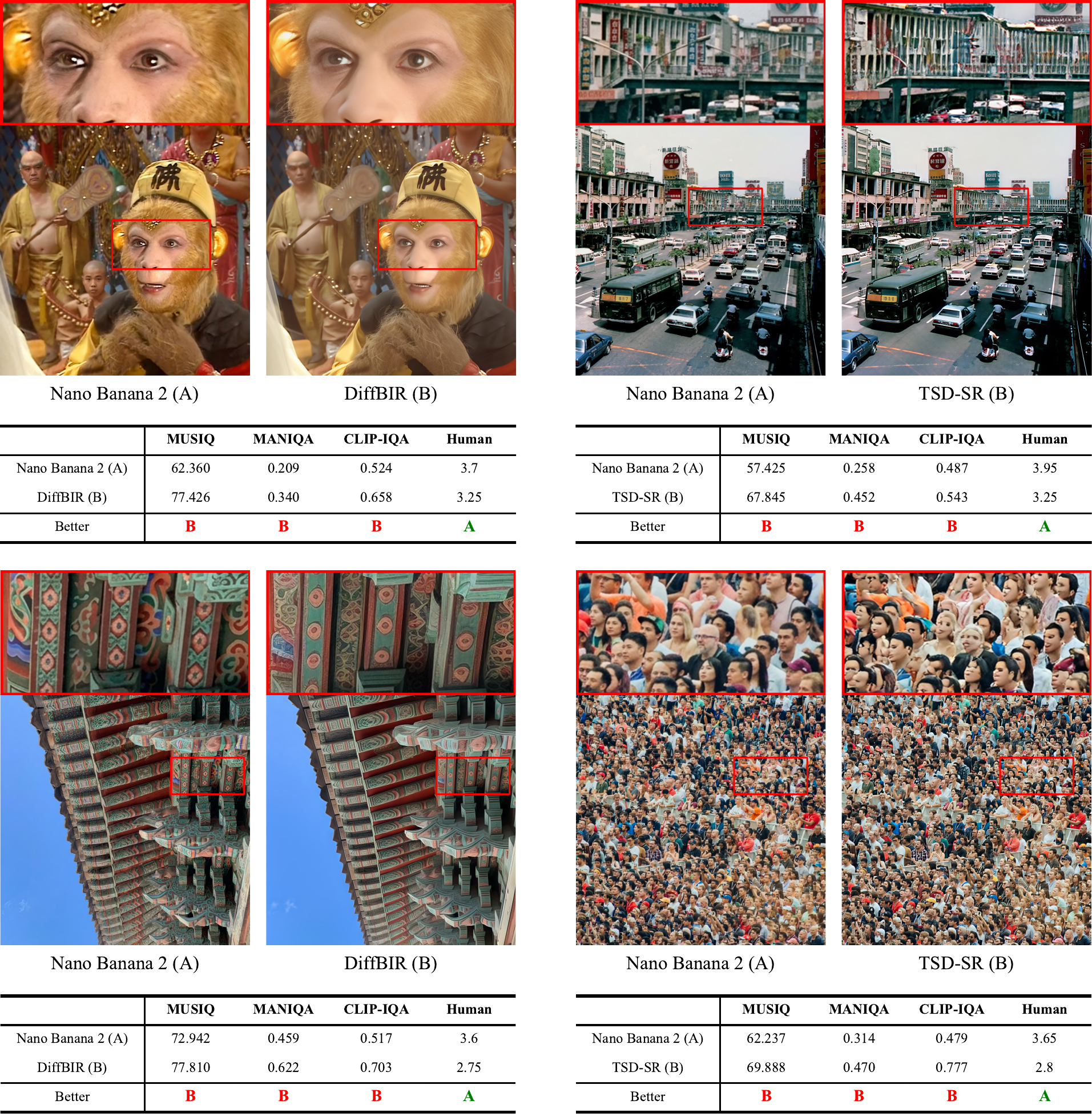} 
    \vspace{-10pt}
    \caption{NR IQA metrics, including MUSIQ, MANIQA, and CLIP-IQA, often fail to align with human perceptual judgments. Each case presents results from two representative methods along with their corresponding NR scores and human evaluation scores. The discrepancy between metric rankings and human preferences is clearly observed across different cases. Zoom in for a better observation.}
    \label{nr_iqa1}
    \vspace{-18pt}
\end{figure*}

\begin{figure*}[tp] 
    \centering
    \includegraphics[width=\textwidth]{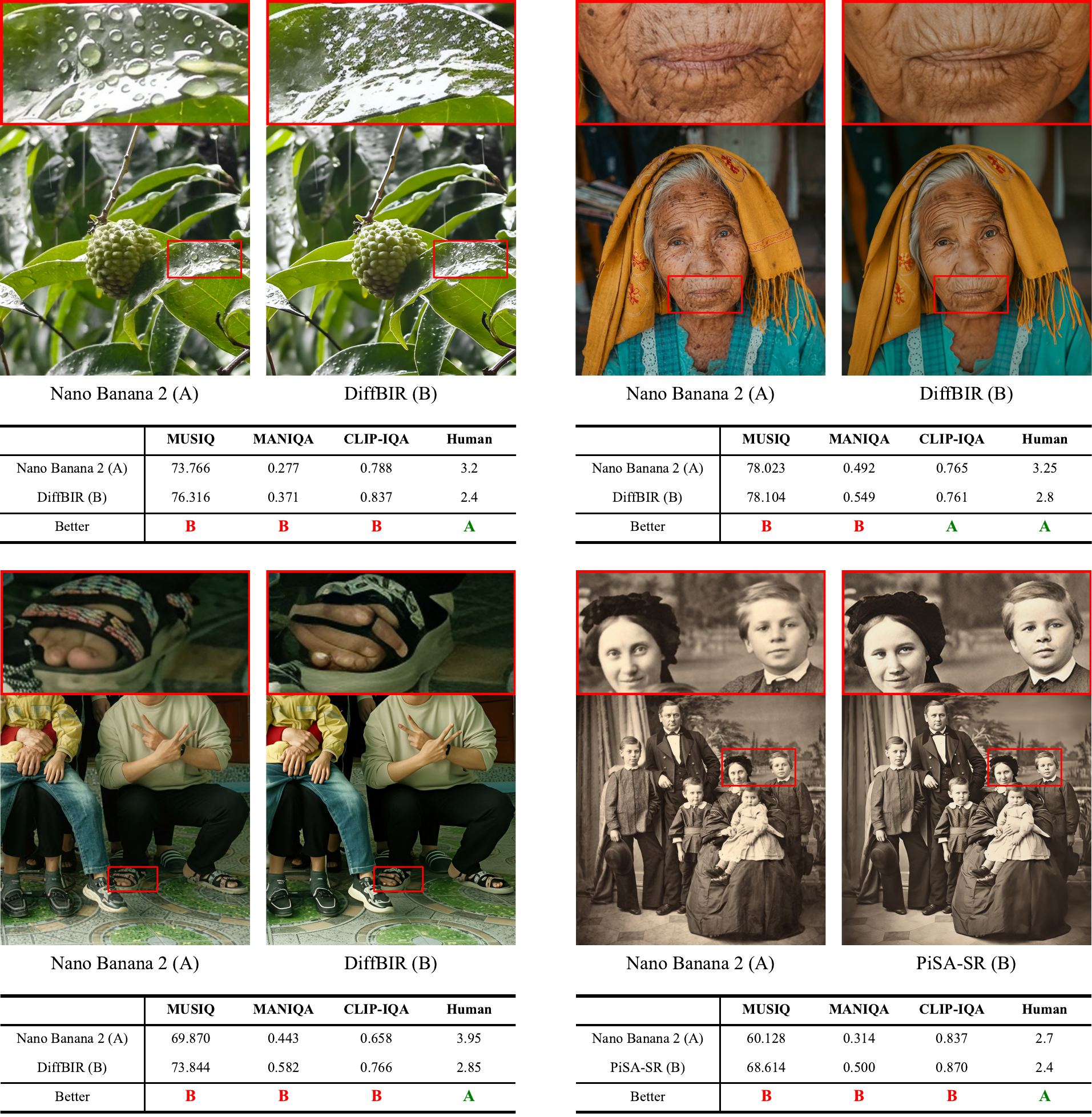} 
    \vspace{-10pt}
    \caption{NR IQA metrics, including MUSIQ, MANIQA, and CLIP-IQA, often fail to align with human perceptual judgments. Each case presents results from two representative methods along with their corresponding NR scores and human evaluation scores. The discrepancy between metric rankings and human preferences is clearly observed across different cases. Zoom in for a better observation.}
    \label{nr_iqa2}
    \vspace{-18pt}
\end{figure*}

In~\cref{nr_iqa1,nr_iqa2}, we further illustrate the limitations of NR IQA metrics, including MUSIQ, MANIQA, and CLIP-IQA. 
Although Nano Banana 2 may occasionally exhibit reduced low-level fidelity compared to reference images, it often produces visually more natural, coherent, and perceptually appealing results from a human perspective. 
However, existing NR IQA metrics frequently fail to capture these perceptual advantages, leading to rankings that are inconsistent with subjective visual quality. 
In several examples, Nano Banana 2 generates realistic textures, harmonious color distributions, and visually pleasing details, yet receives lower NR IQA scores than competing methods that produce oversmoothed results or noticeable artifacts. These discrepancies suggest that current NR IQA metrics remain insufficient for evaluating the perceptual quality of modern generative restoration models.

\begin{figure*}[tp] 
    \centering
    \includegraphics[width=0.995\textwidth]{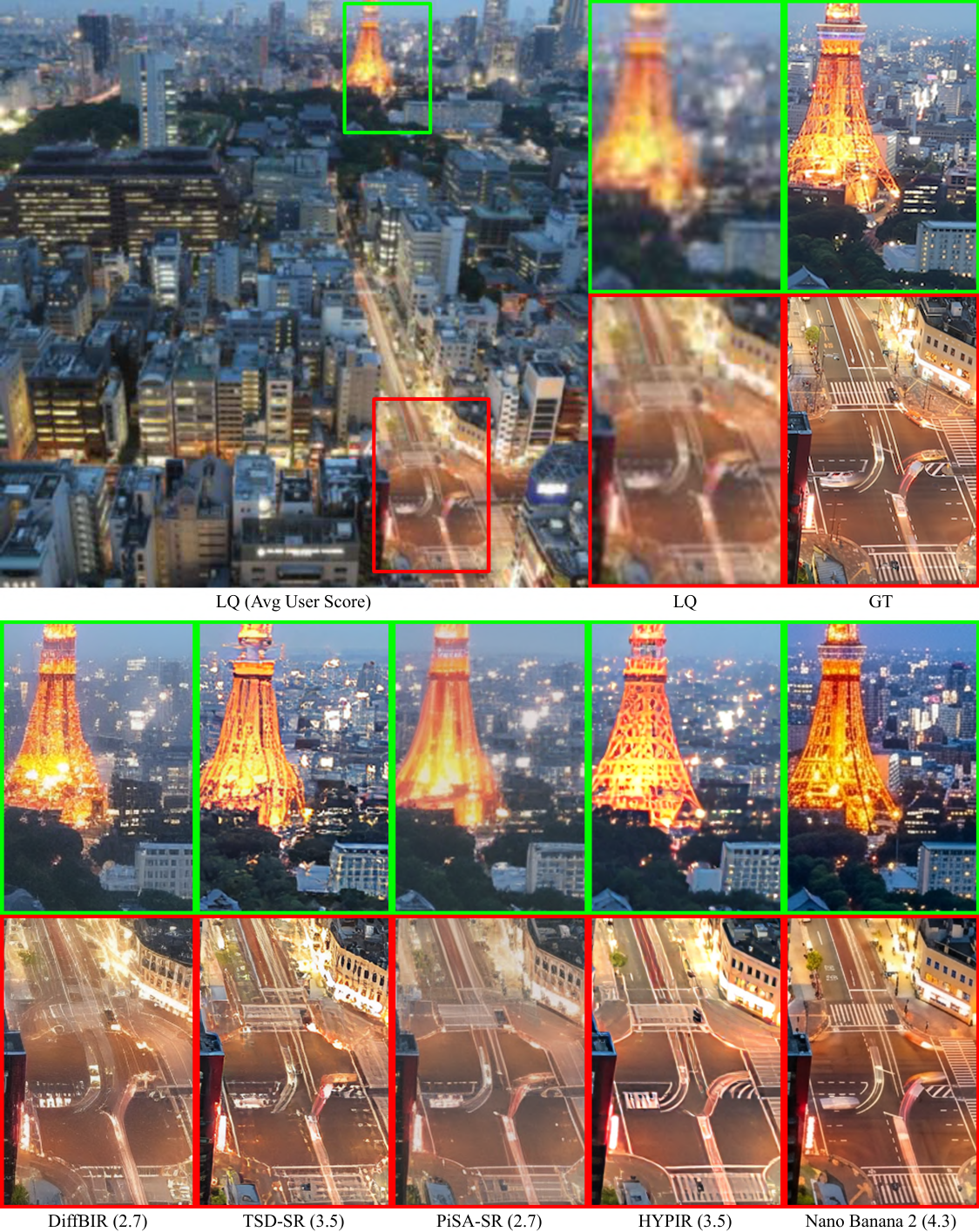} 
    \vspace{-10pt}
    \caption{Our user study adopts a single overall quality score, which is insufficient to fully reflect the trade-off between perceptual quality and restoration fidelity. In the green box, we highlight color inconsistencies in Nano Banana 2, while the red box demonstrates its advantage in restoring fine details in aerial imagery. Zoom in for a better observation.}
    \label{user1}
    \vspace{-18pt}
\end{figure*}

\begin{figure*}[tp] 
    \centering
    \includegraphics[width=0.995\textwidth]{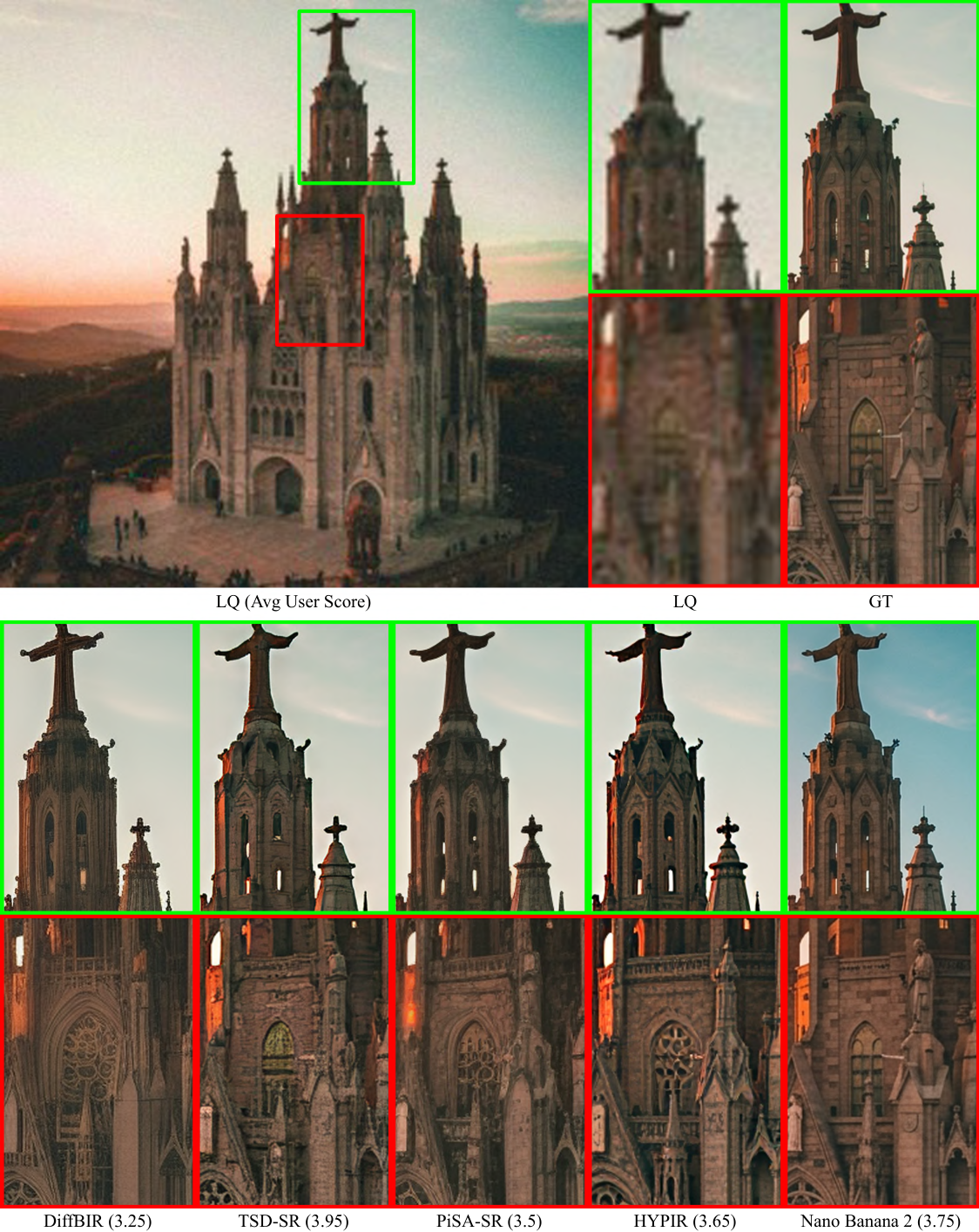} 
    \vspace{-10pt}
    \caption{Our user study adopts a single overall quality score, which is insufficient to fully reflect the trade-off between perceptual quality and restoration fidelity. In the green box, we highlight color inconsistencies in Nano Banana 2, while the red box demonstrates its superiority in recovering fine-grained architectural textures. Zoom in for a better observation.}
    \label{user2}
    \vspace{-18pt}
\end{figure*}

\begin{figure*}[tp] 
    \centering
    \includegraphics[width=0.995\textwidth]{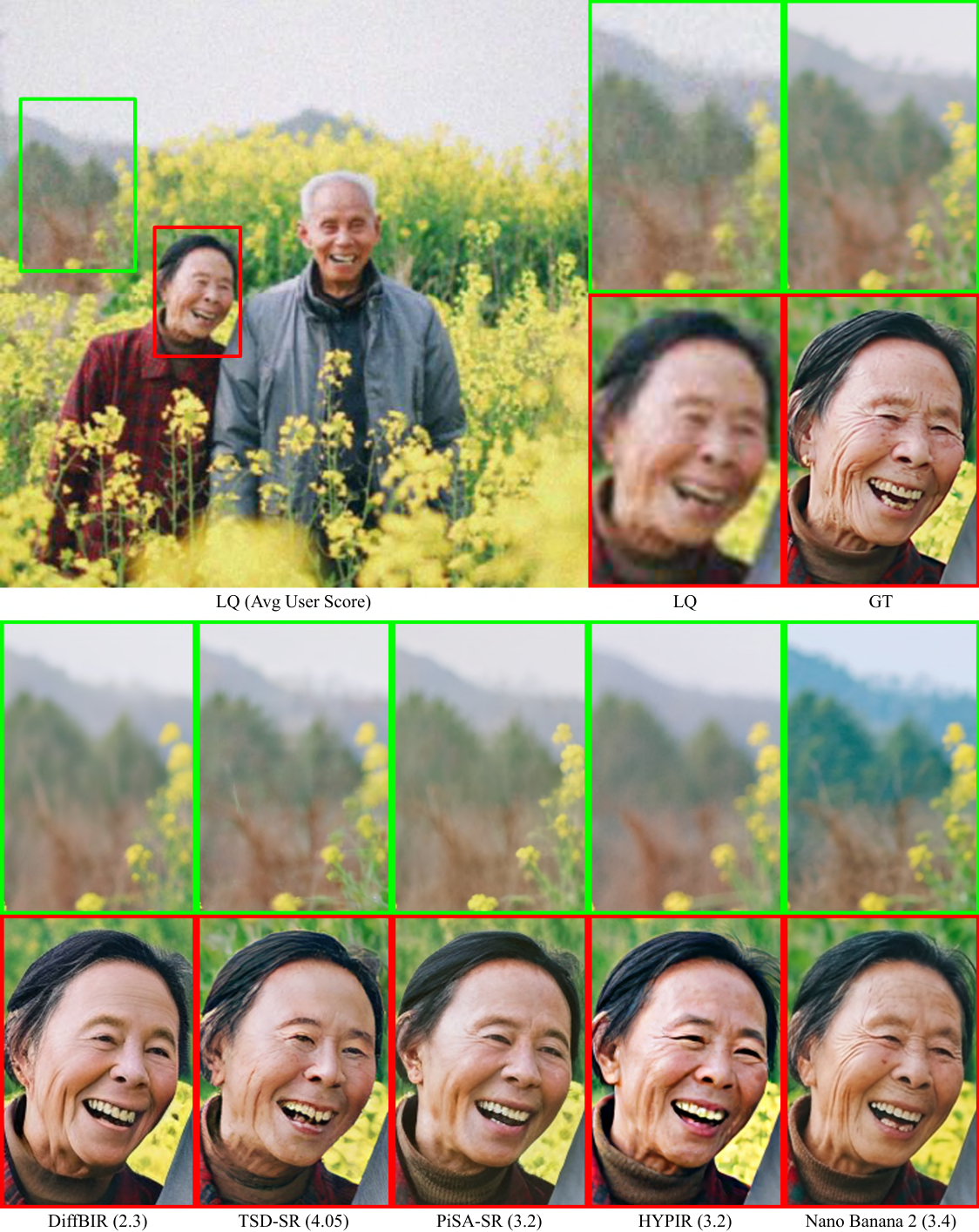} 
    \vspace{-10pt}
    \caption{Our user study adopts a single overall quality score, which is insufficient to fully reflect the trade-off between perceptual quality and restoration fidelity. In the green box, we highlight color inconsistencies in Nano Banana 2, while the red box demonstrates its advantage in restoring facial details. Zoom in for a better observation.}
    \label{user3}
    \vspace{-18pt}
\end{figure*}

\begin{figure*}[tp] 
    \centering
    \includegraphics[width=0.995\textwidth]{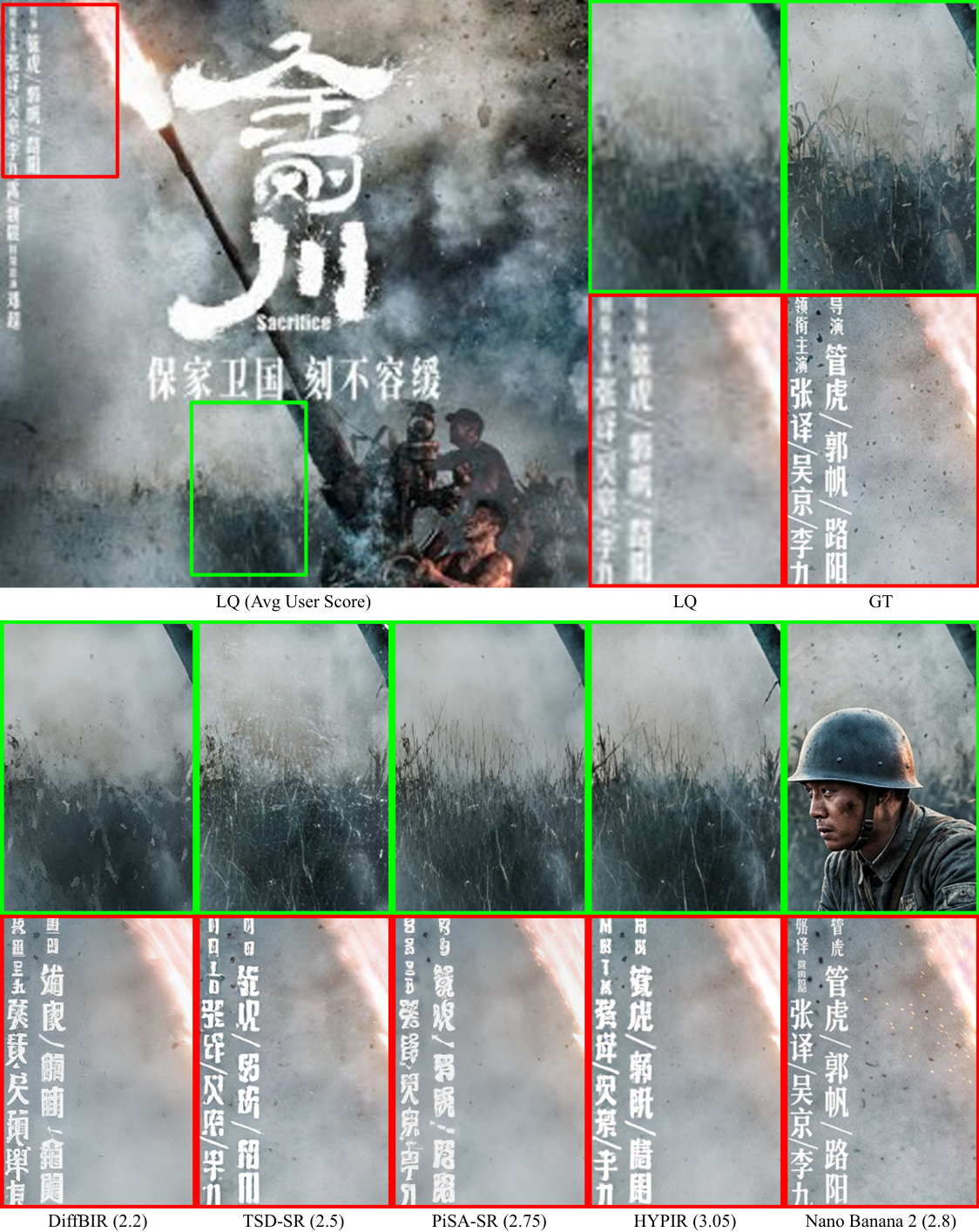} 
    \vspace{-10pt}
    \caption{Our user study adopts a single overall quality score, which is insufficient to fully reflect the trade-off between perceptual quality and restoration fidelity. In the green box, we observe hallucinated content in Nano Banana 2, where an extra person is generated. The red box demonstrates its advantage in text restoration. Zoom in for a better observation.}
    \label{user4}
    \vspace{-18pt}
\end{figure*}

\begin{figure*}[tp] 
    \centering
    \includegraphics[width=0.995\textwidth]{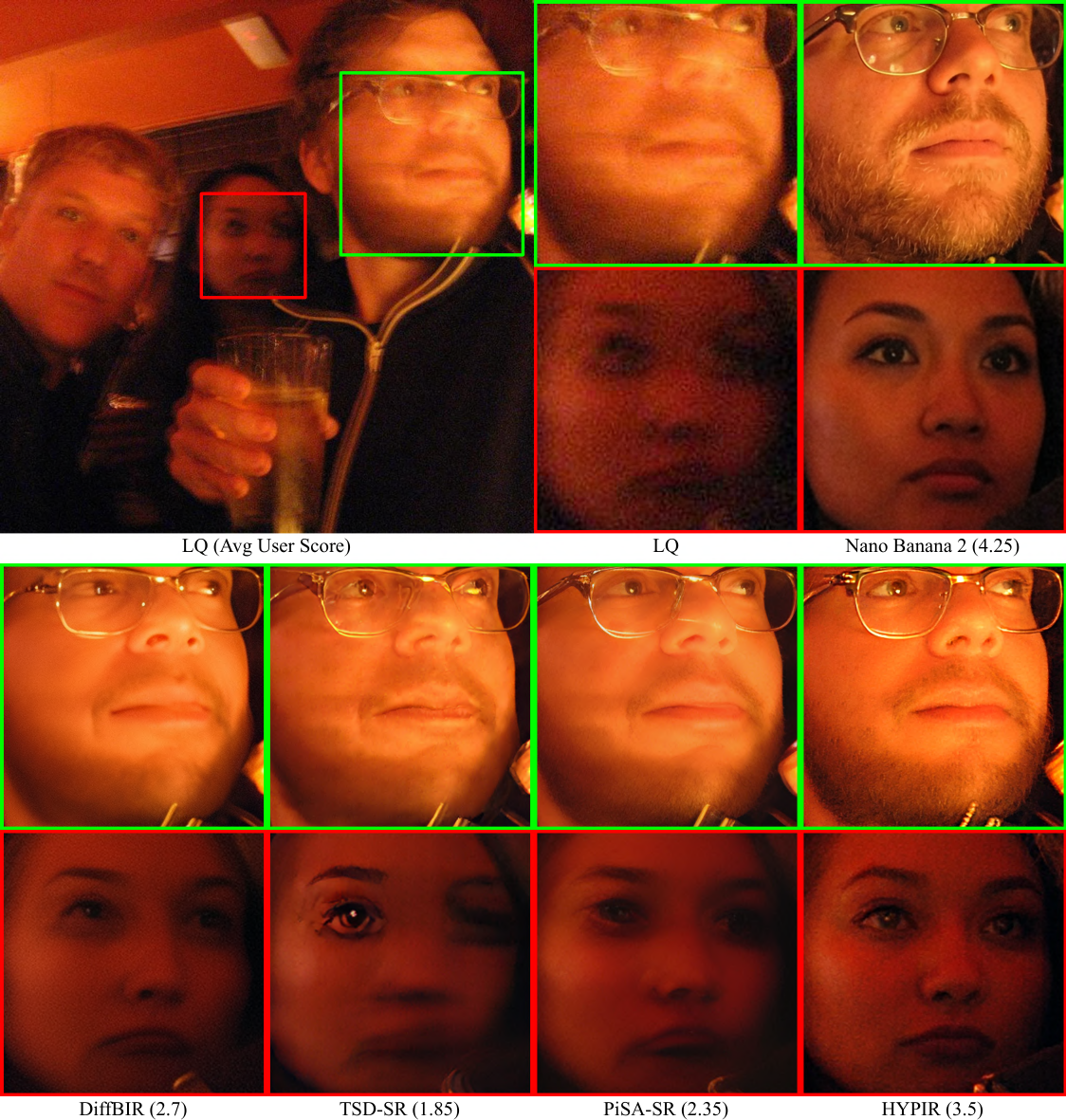} 
    \vspace{-10pt}
    \caption{Our user study adopts a single overall quality score, which is insufficient to fully reflect the trade-off between perceptual quality and restoration fidelity. In the green box, we observe over-generation artifacts in Nano Banana 2, where facial hair is excessively synthesized. The red box demonstrates its advantage in restoring facial details. Zoom in for a better observation.}
    \label{user5}
    \vspace{-18pt}
\end{figure*}

However, we observe that when participants are asked to provide a single overall score during user studies, perceptual quality is often implicitly assigned a higher weight, while fidelity to the input or ground truth may be overlooked. As shown in~\cref{user1,user2,user3,user4,user5}, methods producing visually appealing textures or stronger perceptual enhancement may receive higher user scores despite exhibiting noticeable color deviations, structural inconsistencies, or hallucinated details. Closer inspection reveals discrepancies in low-level fidelity that are not adequately captured by these scores. These observations further suggest that human evaluation of IR results should be conducted from multiple dimensions, jointly considering perceptual quality, structural fidelity, and consistency with the underlying image content~\cite{hu2026position}.

\subsection{Taxonomy of Failure Cases}
\begin{figure*}[tp] 
    \centering
    \includegraphics[width=0.995\textwidth]{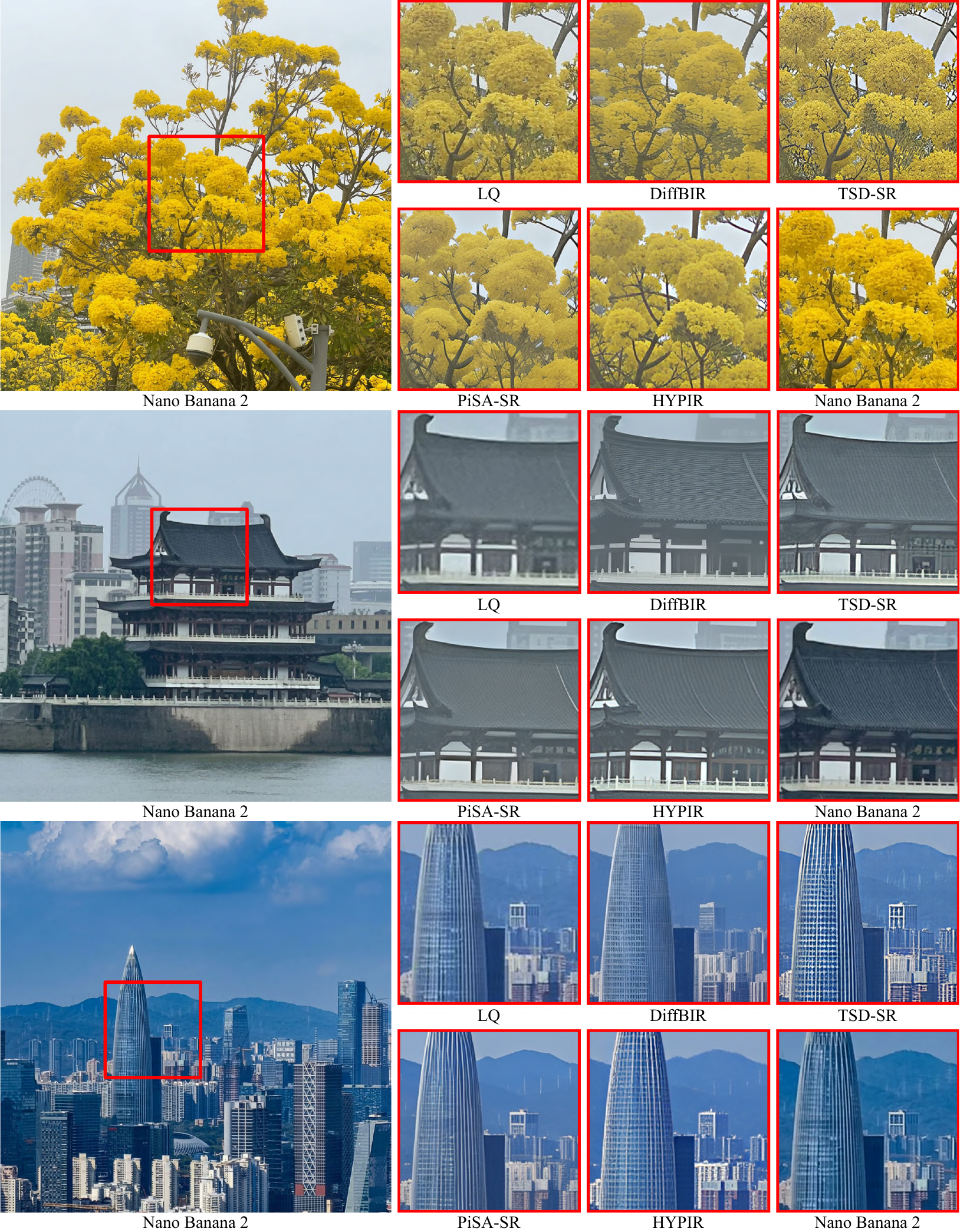} 
    \vspace{-10pt}
    \caption{We observe that Nano Banana 2 may exhibit color and tone inconsistencies, leading to noticeable deviations in hue, saturation, or overall tonal balance despite preserved structural alignment. Zoom in for a better observation.}
    \label{color_1}
    \vspace{-18pt}
\end{figure*}

\begin{figure*}[tp] 
    \centering
    \includegraphics[width=0.995\textwidth]{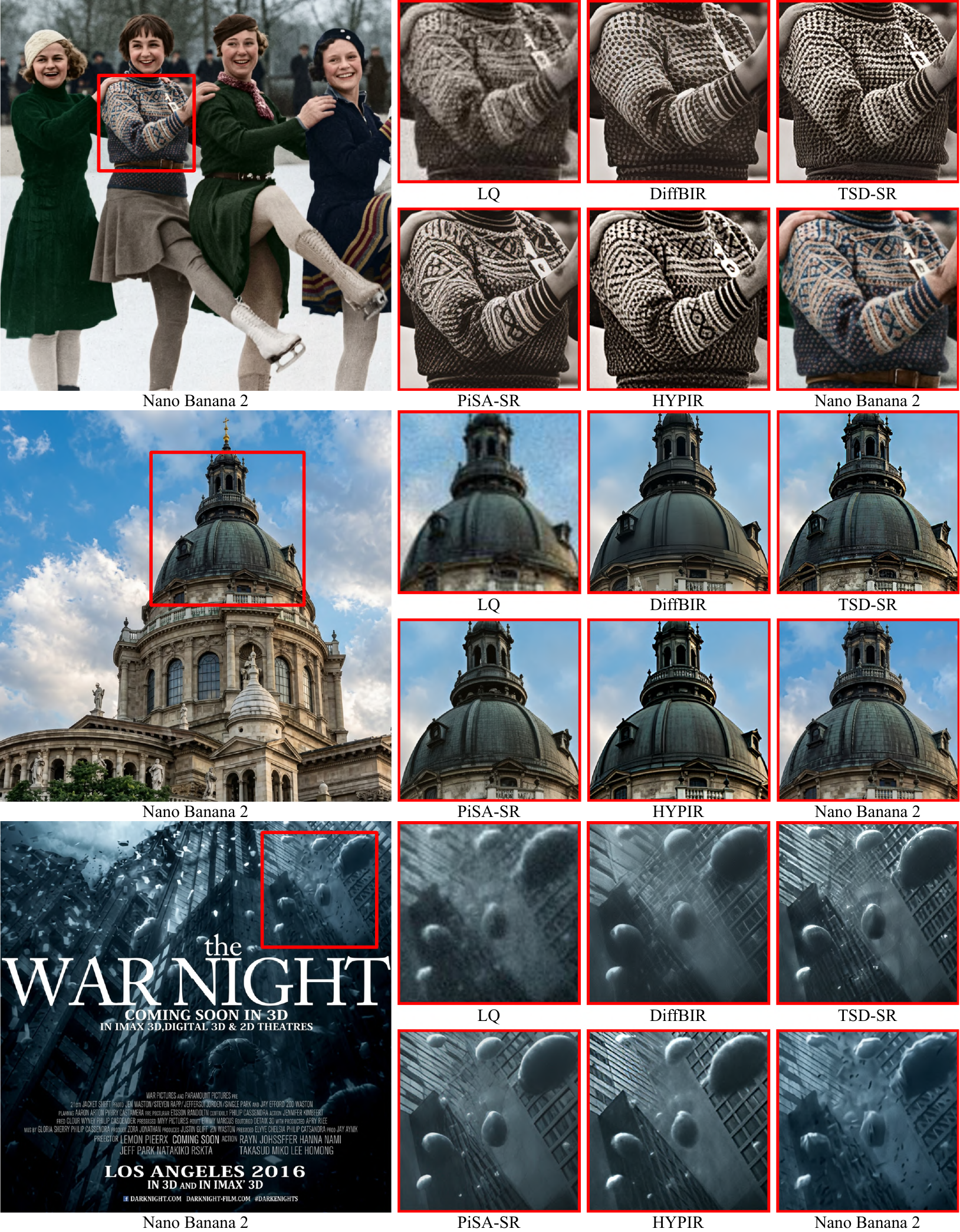} 
    \vspace{-10pt}
    \caption{We observe that Nano Banana 2 may exhibit color and tone inconsistencies, leading to noticeable deviations in hue, saturation, or overall tonal balance despite preserved structural alignment. Zoom in for a better observation.}
    \label{color_2}
    \vspace{-18pt}
\end{figure*}

\begin{figure*}[tp] 
    \centering
    \includegraphics[width=0.995\textwidth]{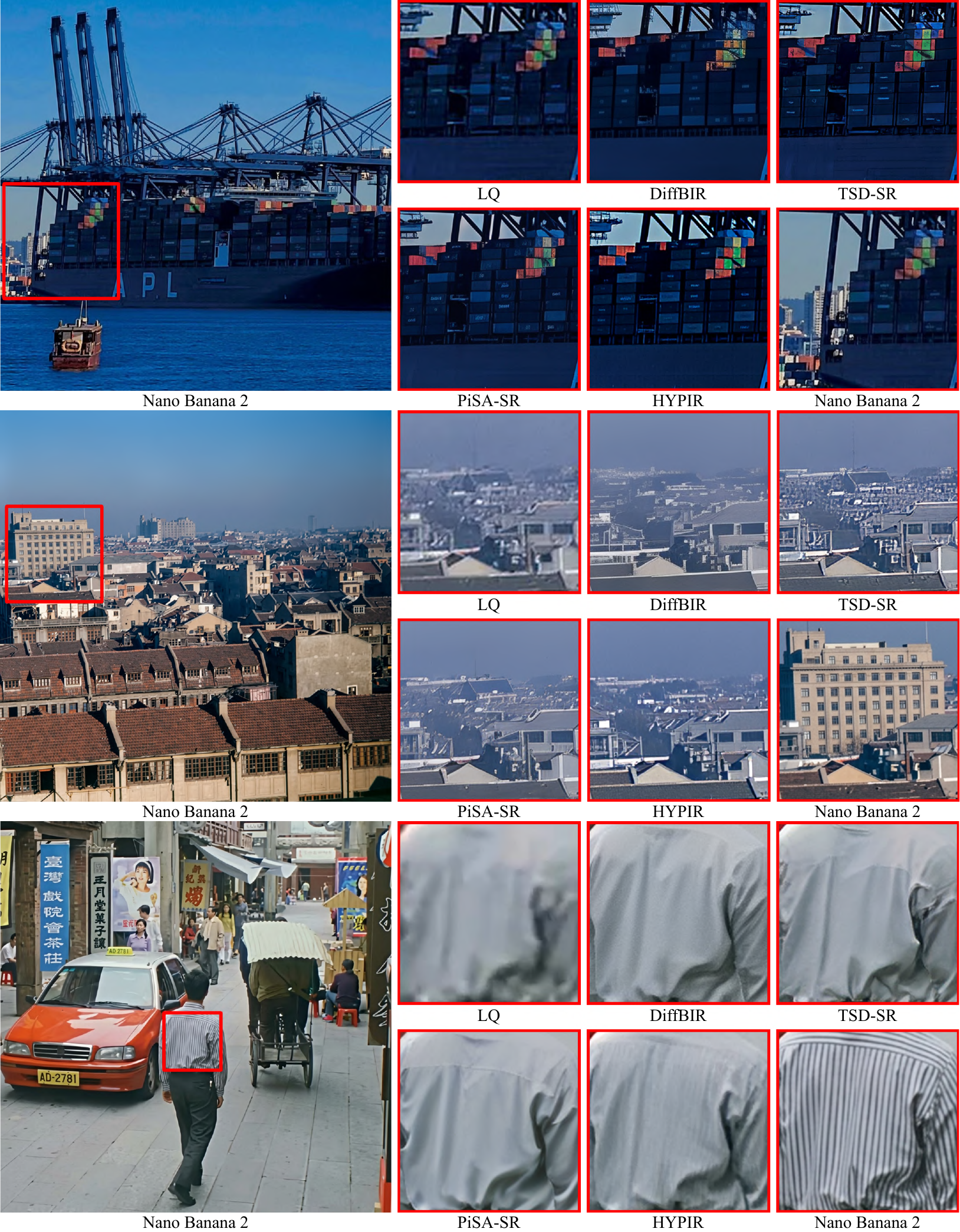} 
    \vspace{-10pt}
    \caption{We observe that Nano Banana 2 may generate hallucinated details, introducing semantically plausible but non-existent content that is not supported by the input. Zoom in for a better observation.}
    \label{hallucination_1}
    \vspace{-18pt}
\end{figure*}

\begin{figure*}[tp] 
    \centering
    \includegraphics[width=0.995\textwidth]{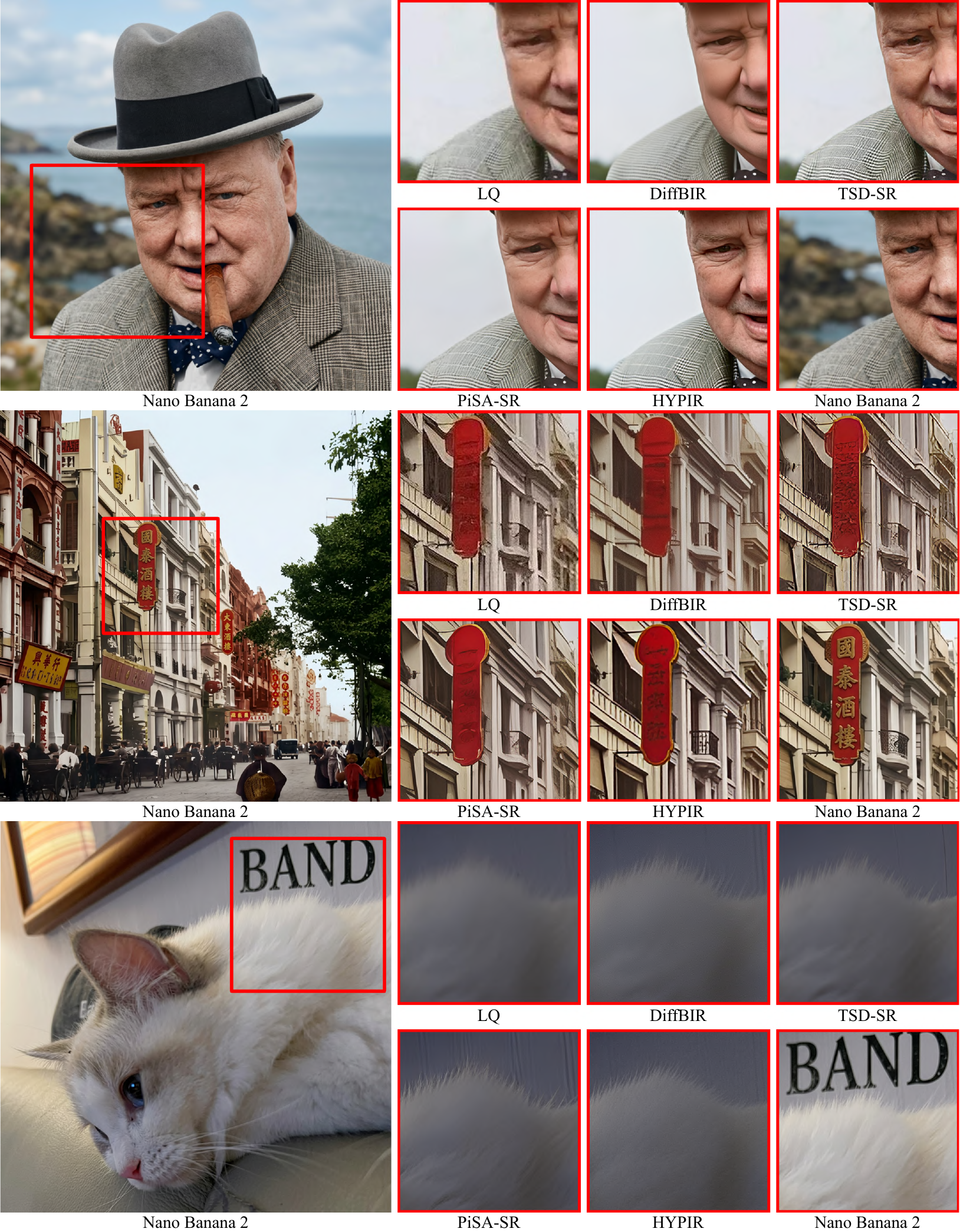} 
    \vspace{-10pt}
    \caption{We observe that Nano Banana 2 may generate hallucinated details, introducing semantically plausible but non-existent content that is not supported by the input. Zoom in for a better observation.}
    \label{hallucination_2}
    \vspace{-18pt}
\end{figure*}

\begin{figure*}[tp] 
    \centering
    \includegraphics[width=\textwidth]{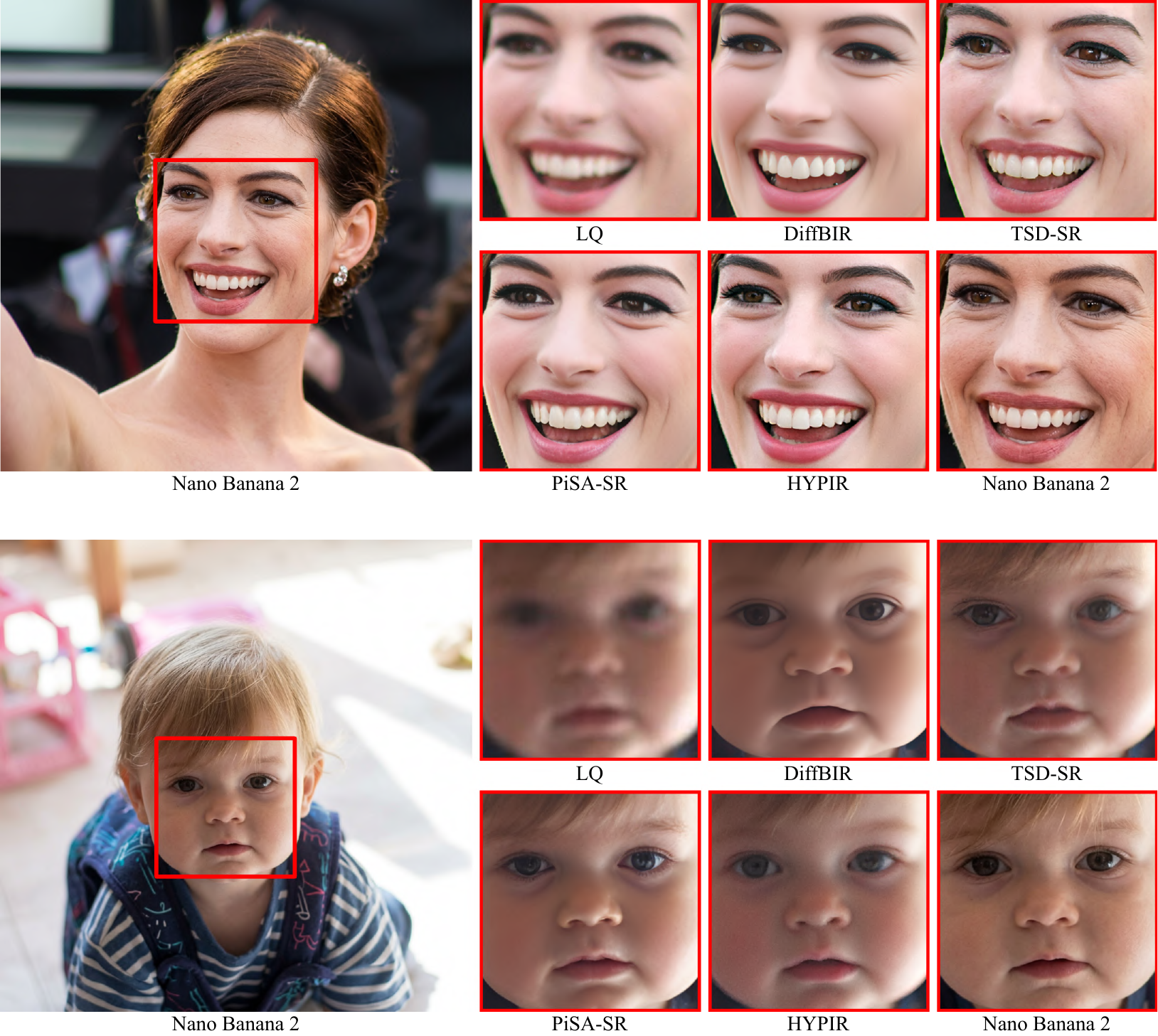} 
    \vspace{-10pt}
    \caption{We observe that Nano Banana 2 may exhibit over-generation, where existing structures or textures are exaggerated beyond what is supported by the input. Zoom in for a better observation.}
    \label{over_generation_1}
    \vspace{-18pt}
\end{figure*}

\begin{figure*}[tp] 
    \centering
    \includegraphics[width=\textwidth]{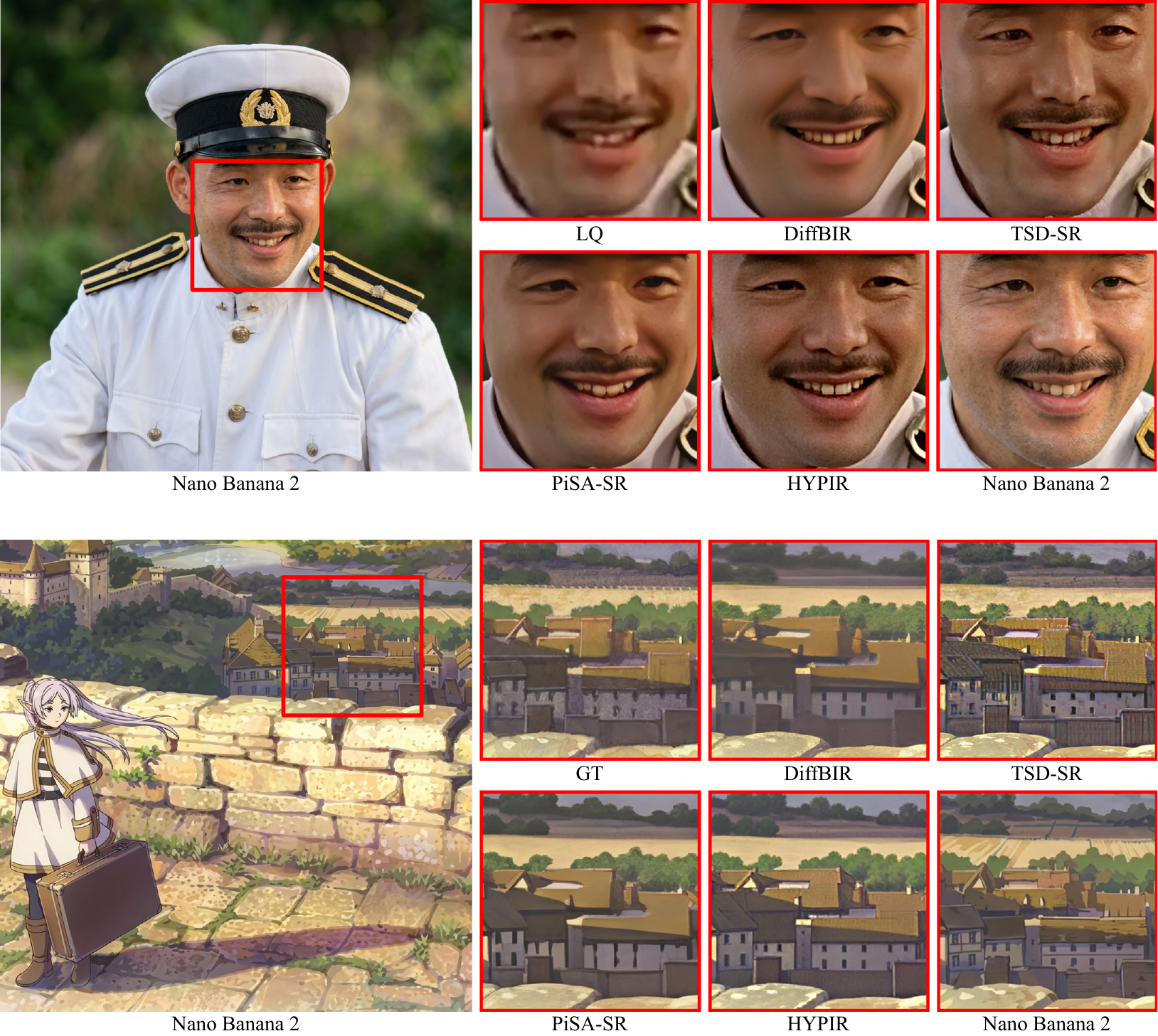} 
    \vspace{-10pt}
    \caption{We observe that Nano Banana 2 may exhibit over-generation, where existing structures or textures are exaggerated beyond what is supported by the input. Zoom in for a better observation.}
    \label{over_generation_2}
    \vspace{-18pt}
\end{figure*}

To provide a more structured analysis beyond quantitative evaluation, we categorize the observed failure cases into three representative types. 
\textbf{(1) Color and tone inconsistencies} refer to deviations in color distribution or tonal balance compared to the input, as shown in~\cref{color_1,color_2}. Even when structural alignment is preserved, the restored results may exhibit noticeable shifts in hue, saturation, or overall appearance.
\textbf{(2) Hallucinated details} refer to the generation of semantically plausible but non-existent content that is not supported by the input, as illustrated in~\cref{hallucination_1,hallucination_2}. This includes introducing new structures, textures, or objects that were not present in the original scene, particularly under severe degradations.
\textbf{(3) Over-generation} refers to the amplification or exaggeration of existing structures or textures in the input, as shown in~\cref{over_generation_1,over_generation_2}. The model enhances high-frequency details beyond what is supported by the degraded observation, leading to overly sharp or dense patterns.
These categories capture distinct failure modes in generative IR, highlighting the gap between perceptual enhancement and faithful reconstruction.

\subsection{Success Cases of Nano Banana 2}

\begin{figure*}[tp] 
    \centering
    \includegraphics[width=\textwidth]{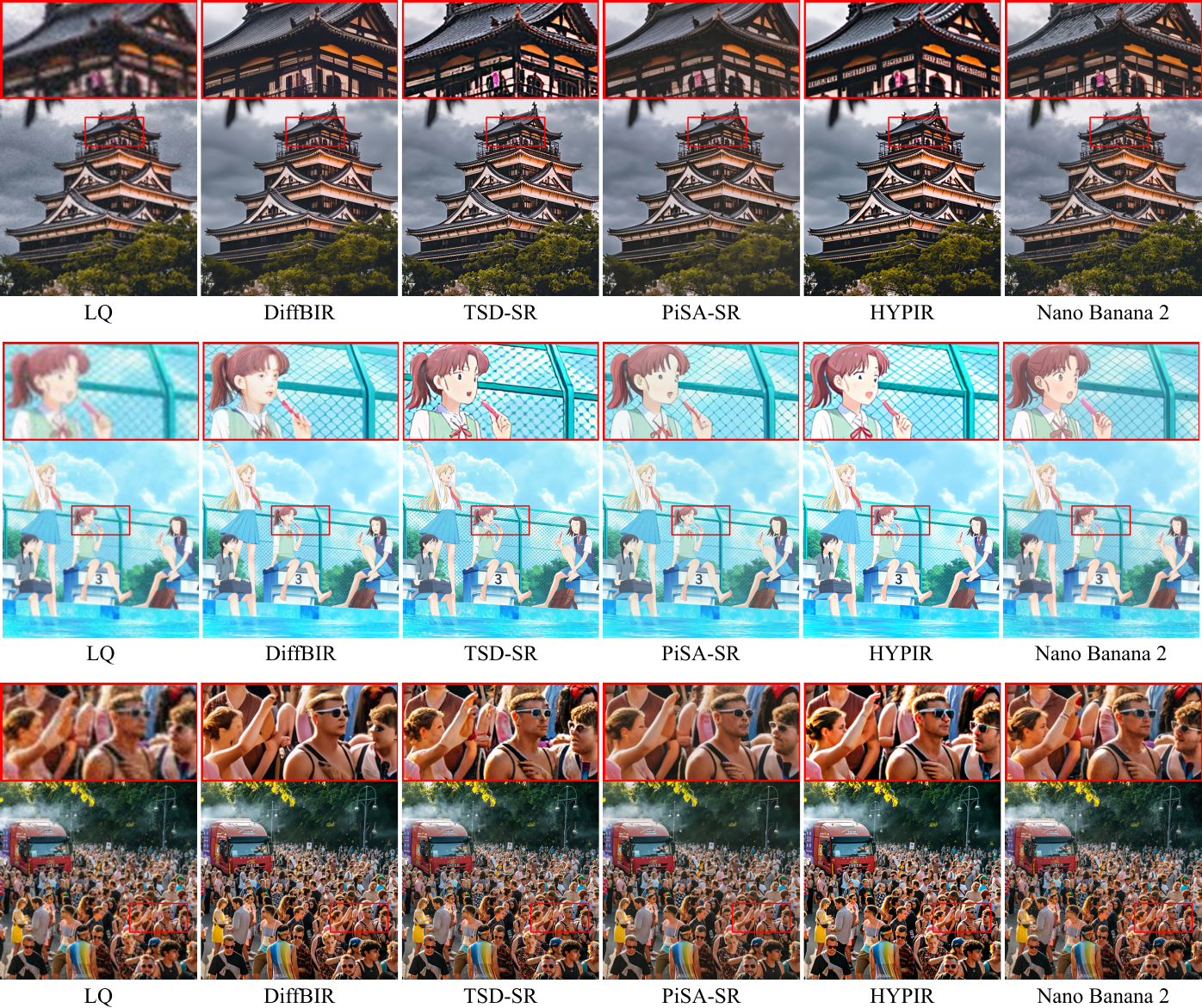} 
    \vspace{-10pt}
    \caption{Representative success cases of Nano Banana 2 compared with previous state-of-the-art methods. Zoom in for a better view.}
    \label{better_1}
    \vspace{-18pt}
\end{figure*}

\begin{figure*}[tp] 
    \centering
    \includegraphics[width=\textwidth]{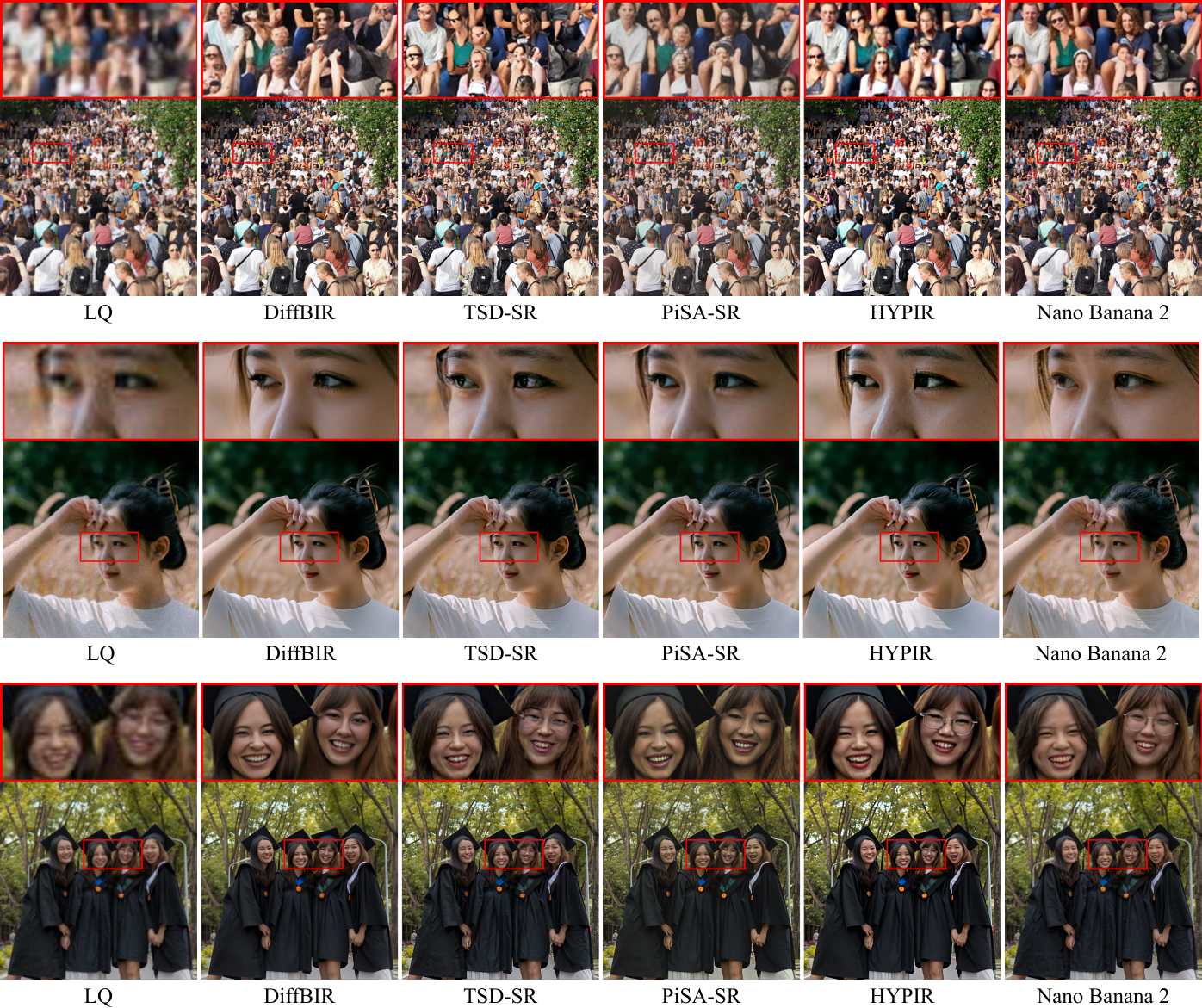} 
    \vspace{-10pt}
    \caption{Representative success cases of Nano Banana 2 compared with previous state-of-the-art methods. Zoom in for a better view.}
    \label{better_2}
    \vspace{-18pt}
\end{figure*}

\begin{figure*}[tp] 
    \centering
    \includegraphics[width=\textwidth]{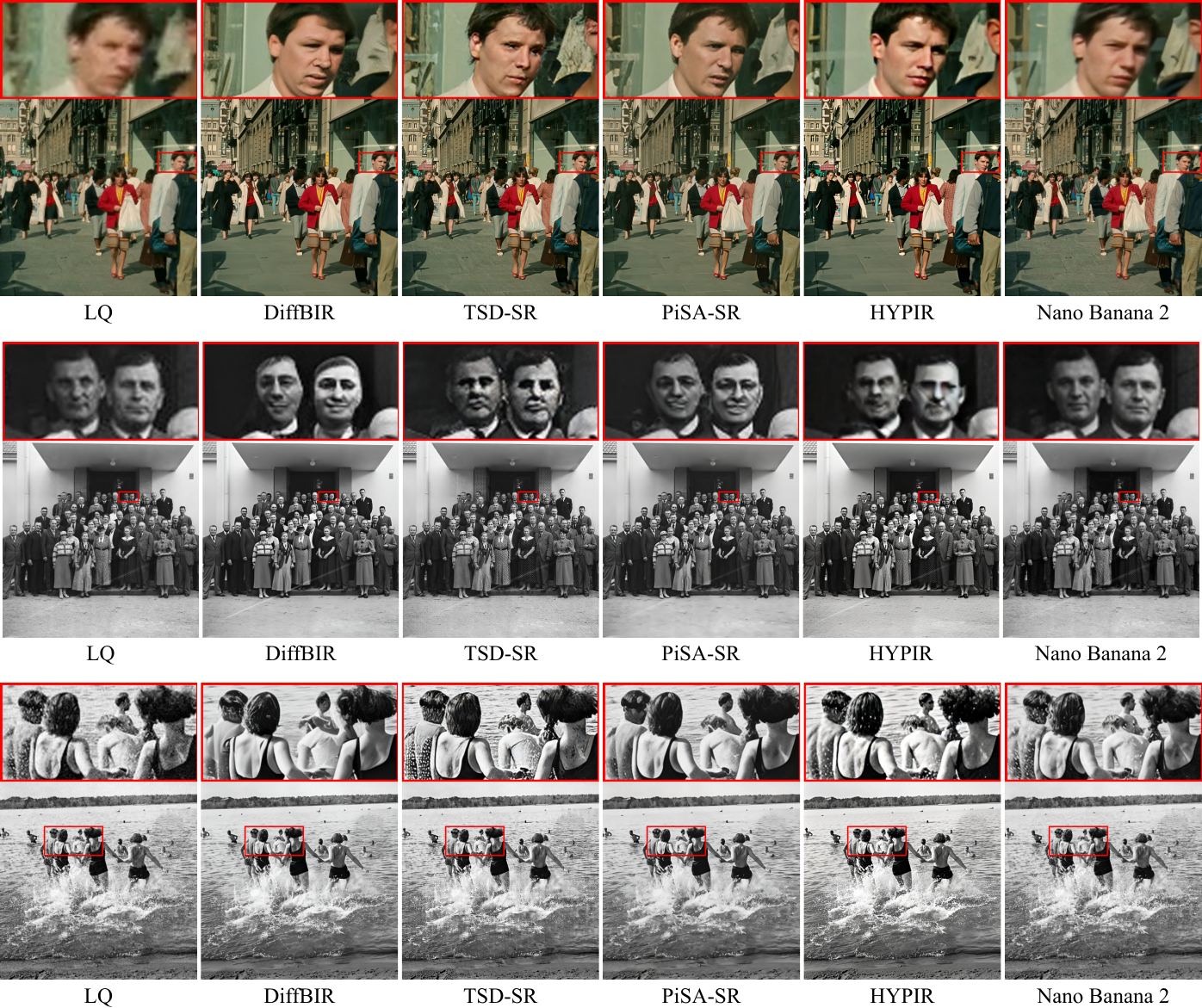} 
    \vspace{-10pt}
    \caption{Representative success cases of Nano Banana 2 compared with previous state-of-the-art methods. Zoom in for a better view.}
    \label{better_3}
    \vspace{-18pt}
\end{figure*}

Beyond the observed failure modes, Nano Banana 2 also demonstrates several notable strengths in challenging restoration scenarios. 
In~\cref{better_1,better_2,better_3}, we present representative success cases where Nano Banana 2 produces more faithful and visually coherent restorations compared with existing state-of-the-art methods. 
Compared with previous generative restoration models, Nano Banana 2 shows a better balance between perceptual enhancement and faithful reconstruction. 
The restored results of Nano Banana 2 are visually sharper and more realistic while simultaneously reducing hallucinated content and over-generation artifacts. 
These examples highlight the potential of Nano Banana 2 to achieve high perceptual quality without substantially sacrificing low-level fidelity, demonstrating its effectiveness across diverse restoration scenarios.

\end{document}